\documentclass[letterpaper]{article}
\usepackage{aaai25} 
\usepackage{times} 
\usepackage{helvet} 
\usepackage{courier}
\usepackage[hyphens]{url}
\usepackage{graphicx}
\usepackage{natbib} 
\usepackage{caption}
\usepackage{algorithm}
\usepackage{algorithmic}
\usepackage{amsmath}

\usepackage{amsmath,amsfonts,bm}

\def\eqref#1{equation~\ref{#1}}

\def\1{\bm{1}}

\DeclareMathAlphabet{\mathsfit}{\encodingdefault}{\sfdefault}{m}{sl}
\SetMathAlphabet{\mathsfit}{bold}{\encodingdefault}{\sfdefault}{bx}{n}

\usepackage{newfloat}
\usepackage{listings}
\usepackage{multirow}
\usepackage{multicol}
\usepackage{xcolor}
\usepackage{colortbl}
\usepackage{makecell}
\usepackage{bm}

\usepackage{microtype}
\DeclareCaptionStyle{ruled}{labelfont=normalfont,labelsep=colon,strut=off}
\floatstyle{ruled}
\newfloat{listing}{tb}{lst}{}
\floatname{listing}{Listing}

\definecolor{custompink}{RGB}{255,105,180}

\title{\textit{Imitate Before Detect}: Aligning Machine Stylistic Preference\\for Machine-Revised Text Detection}
\author{
Jiaqi Chen\textsuperscript{\rm 1,9}\equalcontrib
\quad
Xiaoye Zhu\textsuperscript{\rm 2,10}\equalcontrib 
\quad
Tianyang Liu\textsuperscript{\rm 5}\equalcontrib
\quad
Ying Chen\textsuperscript{\rm 6} \quad
Xinhui Chen\textsuperscript{\rm 3,4} \\
Yiwen Yuan\textsuperscript{\rm 7} \quad
Chak Tou Leong\textsuperscript{\rm 8} \quad
Zuchao Li\textsuperscript{\rm 3}$^\dagger$ \quad
Long Tang\textsuperscript{\rm 12} \quad
Lei Zhang\textsuperscript{\rm 5} \\
Chenyu Yan\textsuperscript{\rm 11} \quad
Guanghao Mei\textsuperscript{\rm 5} \quad
Jie Zhang\textsuperscript{\rm 1}$^\dagger$ \quad
Lefei Zhang\textsuperscript{\rm 3}\thanks{Corresponding author.} \\
}
\affiliations{
    \textsuperscript{\rm 1}Fudan University \quad
\textsuperscript{\rm 2}South China University of Technology  \quad 
\textsuperscript{\rm 3}Wuhan University\\
\textsuperscript{\rm 4}Fenz AI  \quad 
\textsuperscript{\rm 5}UC San Diego  \quad 
\textsuperscript{\rm 6}UIUC  \quad 
\textsuperscript{\rm 7}CMU\\
\textsuperscript{\rm 8}PolyU  \quad 
\textsuperscript{\rm 9}Stanford University  \quad 
\textsuperscript{\rm 10} NUS (Chongqing) Research Institute \\ 
\textsuperscript{\rm 11}Georgia Tech  \quad 
\textsuperscript{\rm 12}Independent researcher \\
\vspace{0.3cm}
\textcolor{custompink}{\texttt{https://machine-text-detection.github.io/ImBD}}
}

\usepackage{bibentry}
\usepackage{booktabs}
\usepackage[standard]{ntheorem}
\newcommand{\note}[1]{{\color{black} #1}}

\begin{document}

\maketitle

\begin{abstract}
Large Language Models (LLMs) have revolutionized text generation, making detecting machine-generated text increasingly challenging. 
Although past methods have achieved good performance on detecting pure machine-generated text, those detectors have poor performance on distinguishing \textbf{\textit{machine-revised text}} (rewriting, expansion, and polishing), which can have only minor changes from its original human prompt.
As the content of text may originate from human prompts, detecting machine-revised text often involves identifying distinctive machine styles, \textit{e.g.}, worded favored by LLMs. 
However, existing methods struggle to detect machine-style phrasing hidden within the content contributed by humans.
We propose the \textbf{\textit{``Imitate Before Detect" (ImBD)}} approach, which first imitates the machine-style token distribution, and then compares the distribution of the text to be tested with the machine-style distribution to determine whether the text has been machine-revised.
To this end, we introduce style preference optimization (SPO), which aligns a scoring LLM model to the preference of text styles generated by machines.
The aligned scoring model is then used to calculate the style-conditional probability curvature (Style-CPC), quantifying the log probability difference between the original and conditionally sampled texts for effective detection.
We conduct extensive comparisons across various scenarios, encompassing text revisions by six LLMs, four distinct text domains, and three machine revision types.
Compared to existing state-of-the-art methods, our method yields a $13$\% increase in AUC for detecting text revised by open-source LLMs, and improves performance by $5$\% and $19$\% for detecting GPT-3.5 and GPT-4o revised text, respectively. 
Notably, our method surpasses the commercially trained GPT-Zero with just $1,000$ samples and five minutes of SPO, demonstrating its efficiency and effectiveness.
\end{abstract}

\section{Introduction}
Large Language Models (LLMs) have demonstrated remarkable capabilities in generating text that is difficult to 
\begin{figure}[!ht]
    \centering
    \includegraphics[width=1.0\linewidth]{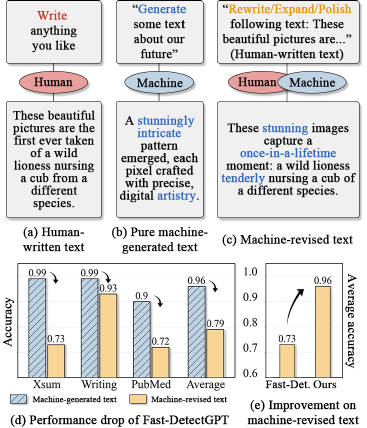}
    \caption{(a-c) Comparative examples of human-written, machine-generated, and machine-revised text. (d) Fast-DetectGPT shows a significant drop in detection accuracy when identifying machine-revised text compared to machine-generated text. (e) Our method brings a noticeable improvement in detecting machine-revised text compared to Fast-DetectGPT. ``Fast-Det." denotes ``Fast-DetectGPT".}
    \vspace{-1em}
    \label{fig:intro}
\end{figure}
distinguish from human writing~\cite{brown2020language,chowdhery2023palm, li2023starcoder, touvron2023llama,touvron2023llama2,chatgpt,openai2024gpt4, deepseekai2024deepseek, lozhkov2024starcoder}. 

With the widespread application of these models, their misuse in exams, academic papers, publications, and other contexts has led to concerns in areas such as academic integrity, fake news, and online information verification. As a result, determining whether a text is LLM-assisted or entirely human-written has become crucial~\cite{bao2023fast}.

In practice, the landscape of LLM-assisted writing extends beyond the widely studied pure generation to also include \textit{\textbf{machine-revised text}}, where LLMs enhance or modify human-written content~\cite{Zhang2024mixtext}.
This shift results in a more nuanced challenge for detection, as the boundaries between human and machine contributions become increasingly intertwined. 
Figure~\ref{fig:intro} (upper) provides comparative examples of human-written, machine-generated text, and machine-revised text. 
This evolution in LLM-assisted writing necessitates a reevaluation of existing detection approaches.
 
Previous detection methods~\cite{hans2024spottingllmsbinocularszeroshot,mitchell2023detectgpt,bao2023fast, su2023detectllm, yang2023dna, zhu2023beat, wu2024wrotethiskeyzeroshot} for identifying machine-generated text rely on calculating classification metrics based on token probabilities from pre-trained language models. 
These methods are built on the assumption that machine-generated texts typically exhibit higher log-likelihoods~\citep{he2024mgtbench, holtzman2020curiouscaseneuraltext} or negative probability curvatures~\citep{mitchell2023detectgpt, bao2023fast} compared to human-written texts. 
While these approaches effectively capture the characteristics of purely machine-generated text, they struggle to identify machine-revised text that contains human content, such as domain-specific terminology. 
This is because the human-contributed content can mislead detectors into believing that the text is human-written~\citep{Zhang2024mixtext,sadasivan2024aigeneratedtextreliablydetected, he2024mgtbench}. 
As a result, these advanced methods experience significant performance drops when detecting machine-revised text (See Figure~\ref{fig:intro} (d)). 
We believe that recognizing the distinctive style of machine-revised text, such as machine-preferred filler phrases and rare vocabulary, is key to effectively detecting such texts.

Specifically, the style distinctions between pure-human and machine-revised texts often lie in subtle stylometric features, as demonstrated by examples in Figure~\ref{fig:intro}. Machine revisions exhibit certain characteristic patterns in word choice (\textit{e.g.}, preference for terms like ``stunning," ``once-in-a-lifetime," and ``tenderly"), sentence structures (\textit{e.g.}, more complex subordinate clauses), and organizational methods (\textit{e.g.}, consistent paragraph structuring)~\cite{chawla2024chatgpt}. 
However, these style features are difficult to capture and isolate due to the human-contributed content mixed into machine-revised text. Therefore, it is necessary to explicitly model these stylistic features.

Motivated by the challenges and observations above, we propose \textbf{\textit{\underline{Im}itate \underline{B}efore \underline{D}etect (ImBD)}} which first imitates the style/pattern of machine-revised texts, then measures the distributional differences between the text under evaluation and the machine style, thereby enabling effective detection of machine-revised texts.
The ImBD consists of two main steps. First, we introduce \textit{\textbf{Style preference optimization (SPO)}} for machine style imitation, which aligns a scoring LLM model to favor the characteristic style of machine-revised text. Specifically, we use pairs of text with identical content – one generated by an LLM and the other written by a human - to adjust the model's token distribution towards a machine-like writing style. Second, we employ the scoring model tuned by step one to calculate the \textit{\textbf{Style-conditional probability curvature (Style-CPC)}}. This metric quantifies the difference between the log probabilities of the original text and alternative versions produced through conditional probability sampling, enabling effective distinction between human-written and machine-revised content. By combining our style-focused alignment with logit-based detection, our method aims to effectively identify machine-revised text even when dealing with advanced language models like GPT-3.5 or GPT-4o.

We demonstrate the efficiency and effectiveness of our method through extensive comparisons across diverse scenarios. 
Our results show significant improvements over existing state-of-the-art methods. We achieve an $13$\% increase in ROAUC for detection on open-source models; $5$\% and $19$\% respective increases on GPT-3.5 and GPT-4o, with limited computational resources – just $1,000$ samples and five minutes of SPO training – our approach outperforms the commercially trained GPT-Zero detector.

Our contributions are three-fold:
\begin{itemize}
    \item  We propose the \textit{\textbf{Imitate Before Detect}} which first imitates the stylistic preferences of LLMs, then measures the distribution distance to recognize machine-revised text that includes human content.
    \item We introduce a comprehensive dataset for machine-revised text detection, enabling robust evaluation of detection methods across diverse domains, revision types, and a wide range of mainstream LLMs.
    \item Our approach achieves $15.16$\%, $19.68$\%, and $12.90$\%  higher ROCAUC than the previous state-of-the-art, Fast-DetectGPT, in detecting revised text from GPT-3.5, GPT-4o, and mainstream open-source LLMs respectively with the same inference speed.
\end{itemize}

\section{Method}
We elaborate on the methods for addressing the challenge of machine-revised text detection, aiming to differentiate between pure human texts and machine-revised texts.
\subsection{Problem Formulation}

Let $x$ denote the given text under detection, represented as a sequence of tokens $\{x_i\}_{i=1}^n$, where $n$ is the length of the sequence. This text $x$ may either be revised by machine or authored by a human. Our primary objective is to utilize a scoring model $p_{\theta}$, which is an autoregressive language model, to ascertain whether the text $x$ is machine-revised ($x_m$) or human-written ($x_h$), thereby formulating this problem as a binary classification task. Formally, we aim to construct a decision function $f: x \rightarrow {0, 1}$, where the output $0$ indicates that the text is human-authored, and $1$ signifies that the text is machine-revised.

\begin{figure}[t]
    \centering
    \includegraphics[width=1.0\linewidth]{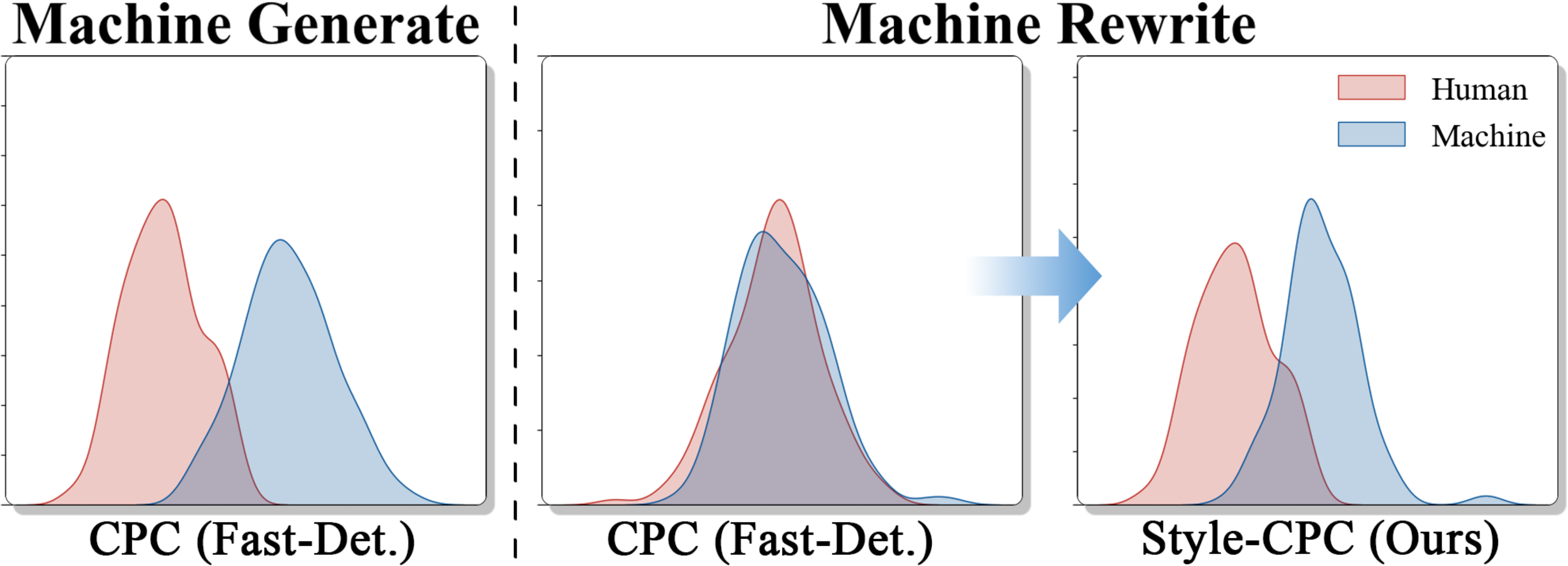}
    \caption{\textbf{Impact of \emph{Style-conditional probability curvatures (Style-CPC).}} (Left) Conditional probability curvatures (CPC) from Fast-DetectGPT (denoted as ``Fast-Det.") applied to purely machine-generated text; (Middle) Conditional probability curvatures applied to purely machine-revised text; (Right) Style-conditional probability curvatures from ours applied to machine-revised text. The greater the separation between human-written texts (red) and machine-revised texts (blue), the more effective the detection.}
    \label{fig:discrepancy-distribution}
\end{figure} 
\subsection{Preliminary}
\paragraph{Foundation} The foundation of machine-generated text detection methods often lies in analyzing the probability distribution of tokens within a given text.
This is rooted in the fact that common decoding strategies, such as top-k, top-p, and beam search, favor high-likelihood next tokens in autoregressive generation, while high-quality human language does not necessarily follow high-probability next words \cite{holtzman2020curiouscaseneuraltext}.

To quantify the differences between machine-generated text \(x_m\) and human-written text \(x_h\), one effective strategy is to measure the discrepancy ($\delta$) between the log probability of the original text and its alternative versions under perturbation~\cite{mitchell2023detectgpt} or after resampling~\cite{bao2023fast}. Let \(\phi\) denote a transformation function that produces an altered version \(\tilde{x}\) from the original text \(x\), \textit{i.e.}, \(\tilde{x} \sim \phi(x)\). In machine-generated texts, the original tokens often have higher probabilities, and after applying \(\phi\) for token replacement, the probabilities of the new tokens tend to be lower on average. Conversely, human-written texts typically exhibit a more diverse range of token probabilities, leading to a smaller discrepancy after alterations. As a result, this discrepancy tends to be larger for machine-generated text compared to human-written text. Formally, we can express this inequality as:
\definecolor{lightblue}{RGB}{235,245,250}
\definecolor{lightmint}{RGB}{255,239,239}
\definecolor{darkblue}{RGB}{0,90,120}
\definecolor{darkgreen}{RGB}{120,0,0}

\begin{equation*}
\begin{gathered}
\textcolor{darkblue}{
  \underbrace{
    \colorbox{lightblue}{
      \color{black}$\log p(x_m)-\mathbb{E}_{\tilde{x}_m \sim \phi(x_m)}\log p(\tilde{x}_m)$
    }
  }_{\text{discrepancy of machine-generated text ($\delta_m$)}}
}
\\
\text{\large $>$ }
\textcolor{darkgreen}{
  \underbrace{
    \colorbox{lightmint}{
      \color{black}$\log p(x_h)-\mathbb{E}_{\tilde{x}_h \sim \phi(x_h)}\log p(\tilde{x}_h)$
    }
  }_{\text{discrepancy of human-written text ($\delta_h$)}}
}
\end{gathered}
\end{equation*}

where $p$ represents the probability distribution of the source model. The source model can be effectively replaced by a substitute scoring model $p_\theta$ in black-box scenarios \cite{mitchell2023detectgpt}. This inequality forms the basis for distinguishing between machine-generated and human-written content. Recent studies have demonstrated the effectiveness of this approach in detecting machine-generated text \cite{mitchell2023detectgpt,bao2023fast}. In scenarios where the distributions of these discrepancies show a small overlap between machine-generated and human-written texts, this approach can effectively distinguish between the two types of content. As shown in Figure~\ref{fig:discrepancy-distribution} (left), the distribution of the discrepancy for machine-generated text is generally larger than that for human-written text, creating a gap that allows differentiation between the two.

\paragraph{Problem Analysis}
While the aforementioned approach can be effective for detecting pure machine-generated text, it encounters significant challenges when applied to more nuanced scenarios, particularly in the detection of machine-revised texts. 
In tasks, such as \texttt{rewrite} or \texttt{polish}, where machines make small changes on top of human writing, we observe a substantial overlap in the probability distributions of machine-revised and human-written texts, as shown in Figure~\ref{fig:discrepancy-distribution} (right).

\begin{figure}[!t]
    \centering
    \includegraphics[width=1.0\linewidth]{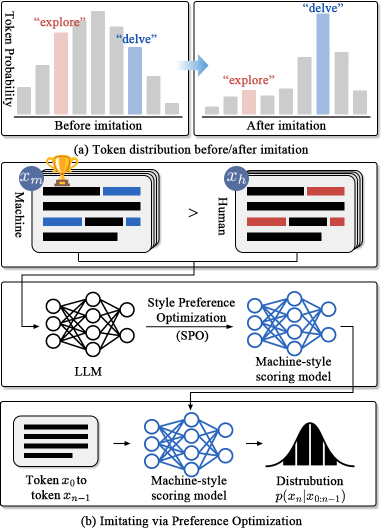}
    \caption{\textbf{Imitating the stylistic preferences of LLMs.} (a) Token distribution before and after machine-style imitation, demonstrating a deliberate fine-tuning of the scoring model to bias its token distribution towards a machine writing style (\textit{e.g.}, shifting preferences from common words like ``explore" to machine-favored tokens such as ``delve''). (b) The pipeline of Style Preference Optimization is applied to align the base scoring model with the style of machine-revised content using paired human-machine texts. This results in a machine-style scoring model, which generates token distributions $p(x_n|x_{0:n-1})$ for each position \( n \), subsequently used for style-conditional probability curvature calculations.}
    \label{fig:detection-process}
\end{figure}
This overlap severely compromises the effectiveness of detection methods that rely on the hypothesis. The limitations arise from two key factors.
First, when users provide part of the content, the resulting text is not entirely ``machine-generated", making probability-based distinctions less effective. Second, advanced LLMs may develop unique stylistic patterns that are not captured by traditional methods. 
For instance, models like GPT-4 might favor words such as \textit{commendable}, ``\textit{embark}", ``\textit{delve into}", ``\textit{intricate}", \textit{etc.}~\cite{liang2024monitoringaimodifiedcontentscale, gray2024chatgptcontaminationestimatingprevalence, chawla2024chatgpt}, in contexts where a scoring model trained on a general corpus would consider them unexpected. 
This discrepancy skews the calculation of the probability curvature, leading to values that significantly overlap between machine-revised and human-written texts, making reliable distinction challenging.

These challenges underscore the need for a more nuanced approach to detection that focuses on capturing the subtle stylistic differences between human-written and machine-revised text. 
Therefore, we propose to learn the characteristic style of machine-revised text by imitating the token distribution output by LLMs. 
By focusing on style rather than content, we aim to enhance the detector's ability to distinguish between human-written and machine-revised text.

\subsection{Imitating via Preference Optimization}
Based on the challenges identified in detecting machine-revised text, we observed that the key to effective detection lies in increasing the discrepancy between the probability distributions of machine-revised and human-written texts. To address this, we aim to increase the difference between the discrepancies $\delta_m$ and $\delta_h$, as defined earlier. 
Specifically, our objective is to optimize the scoring model $p_\theta$ to better imitate the token distribution with machine style, such that:

$$
\max_{p_\theta} \mathbb{E}_{x_m, x_h} [\delta_m - \delta_h].
$$

This objective seeks to widen the gap between the discrepancies between machine-revised and human-written texts, making them more distinguishable. To achieve this, we propose a method called style preference optimization, which leverages preference learning to tune the scoring model $p_\theta$ towards favoring machine-revised text patterns.

As shown in Figure~\ref{fig:detection-process} (b), the core of this method involves constructing preference relations between pairs of texts with equivalent content: one human-written ($x_h$) and one machine-revised ($x_m$). These pairs are created through a rewriting process, ensuring that the content remains consistent while the writing style varies. This pairing strategy allows us to isolate and focus on stylistic differences, controlling for content variability. By optimizing the scoring model $p_\theta$ to exhibit a stronger preference for the stylistic features of machine-revised text $x_m$ over those of human-written text $x_h$, we denote this preference as $x_m \succ x_h$. We formulate this preference learning through the lens of reward learning. Assuming an optimal reward function $r$, we express the preference distribution $p^*$ using the Bradley-Terry model:
$$
p^*(x_m \succ x_h) = \sigma(r(x_m) - r(x_h)),
$$
where $\sigma$ is the sigmoid function. This formulation indicates that the probability of preferring machine-revised text over human-written text increases as the reward difference $r(x_m) - r(x_h)$ grows. Following the Direct Preference Optimization (DPO) approach, we reparameterize the reward function $r$ using a closed-form expression based on the optimal policy:
$$
r(x) = \beta \log \frac{p_\theta(x)}{p_{\theta_{\text{ref}}(x)}}.
$$
Here, $p_{\theta_{\text{ref}}}$ represents a reference model, typically the initial state of $p_\theta$ before optimization. By incorporating this reward formulation, we express the probability of preference data with the policy model rather than the reward model. Given a training dataset $\mathcal{D}$ of content-equivalent $(x_m, x_h)$ pairs, we optimize the following objective:
$$
\underset{p_\theta}{\mathrm{max}}\ \mathop{\mathbb{E}}_{(x_m, x_h) \sim \mathcal{D}}\ \left[ \log \sigma \left( \beta \log \frac{p_\theta(x_m)}{p_{\theta_{\text{ref}}(x_m)}} - \beta \log \frac{p_\theta(x_h)}{p_{\theta_{\text{ref}}(x_h)}} \right) \right].
$$
By optimizing this objective function, we can adjust the model $p_\theta$ to favor the stylistic features of machine-revised texts. This adjustment makes the model more sensitive to the stylistic characteristics of machine-revised text. We denote the optimized model as $\hat{p_\theta}$, representing a machine-style scoring model that is strongly more aligned with machine-revised text styles.

\subsection{Detection via Style Probability Curvature}
After aligning our model with machine-revised text styles, we proceed with the detection step using conditional probability curvature~\cite{bao2023fast}. Specifically, given the machine-style scoring model $\hat{p_\theta}$ and a sampling model $q_\phi$, we define the style-conditional probability as:
$$
p(\tilde{x}|x) = \prod_{j} \hat{p_\theta}(\tilde{x}_j | x_{<j}).
$$
Here, \(\tilde{x}\) is generated by sampling each token \(x_i\) from \(\hat{p_\theta}(x_i \mid x_{<i})\) without conditioning on other sampled tokens. The \textit{style-conditional probability curvature (Style-CPC)} is quantified as:
$$
\mathbf{d}(x, \hat{p_\theta}, q_\phi) = \frac{\log \hat{p_\theta}(x | x) - \tilde{\mu}}{\tilde{\sigma}},
$$
where
$$
\tilde{\mu} =  \mathbb{E}_{\Tilde{x} \sim q_\phi  (\Tilde{x} \mid x)} \left( \log p_\theta(\tilde{x}_i\mid  x) \right)\quad,
$$
$$
\quad \tilde{\sigma}^2 =  \mathbb{E}_{\Tilde{x} \sim q_\phi (\Tilde{x} \mid x)}\left( \log p_\theta(\tilde{x}_i\mid  x)  - \tilde{\mu}^2 \right).
$$
This metric $\mathbf{d}(x, \hat{p_\theta}, q_\phi)$ allows us to quantify the log probability difference between the original and alternative sampled texts. 
Figure~\ref{fig:discrepancy-distribution} illustrates the distribution of $\mathbf{d}$ before and after applying Style-CPC. We observe that using the aligned model to calculate $\mathbf{d}$ significantly reduces the overlap between distributions of human-written and machine-revised texts. This reduced overlap enables us to identify an effective threshold value $\epsilon$, leading to a straightforward classification strategy:
$$
f(x) = \begin{cases} 
1 & \text{if } \mathbf{d}(x, \hat{p_\theta}, q_\phi) > \epsilon \\
0 & \text{otherwise}
\end{cases},
$$
where $f(x) = 1$ indicates machine-revised text, and $f(x) = 0$ signifies human-written text.
By combining machine style alignment with probability curvature detection, our method aims to enhance the model's sensitivity to the unique stylistic features of machine-revised texts. Essentially, we tune the scoring model to be more biased towards machine-revised styles, making it `aware' of the subtle differences between machine and human writing styles. This increased sensitivity allows for a more pronounced separation in the probability curvature distributions of machine and human-authored texts. Consequently, the previously overlapping distributions become more distinct, enabling effective logits-based detection that was previously challenging. This approach shifts the focus from content to style, seeking to address the limitations of traditional methods in detecting outputs from advanced language models and in scenarios with user-provided content.

\section{Experiment}
\begin{table*}[t]

\small
\centering
\setlength{\tabcolsep}{11pt}
\renewcommand\arraystretch{1.2}
    \begin{tabular}{l|c|@{\hspace{11.5pt}}c@{\hspace{11.5pt}}c@{\hspace{11.5pt}}c@{\hspace{11.5pt}}c|@{\hspace{11.5pt}}c@{\hspace{11.5pt}}c@{\hspace{11.5pt}}c@{\hspace{11.5pt}}c@{\hspace{11.5pt}}c}
\hline

\hline

\hline

\hline
\multirow{2}{*}{\bf Method} & \bf Time cost & \multicolumn{3}{c}{\bf GPT-3.5} & \multirow{2}{*}{\bf Avg.} & \multicolumn{3}{c}{\bf GPT-4o} & \multirow{2}{*}{\bf Avg.} \\
&(s/$1$k words) & XSum & Writing & PubMed & & XSum & Writing & PubMed & \\

\hline

\hline
        RoBERTa-base & \bf{\color[HTML]{000000} \bm{$0.07$}} & {\color[HTML]{000000} $0.5806$} & {\color[HTML]{000000} $0.7225$}& {\color[HTML]{000000} $0.4370$} & {\color[HTML]{000000} $0.5800$}  & {\color[HTML]{000000} $0.4921$} & {\color[HTML]{000000} $0.4774$} & {\color[HTML]{000000} $0.2496$} & {\color[HTML]{000000} $0.4064$} \\
        RoBERTa-large & {\color[HTML]{000000} $0.11$} & {\color[HTML]{000000} $0.6391$} & {\color[HTML]{000000} $0.7236$} & {\color[HTML]{000000} $0.4848$} & {\color[HTML]{000000} $0.6158$} & {\color[HTML]{000000} $0.4782$} & {\color[HTML]{000000} $0.4708$} & {\color[HTML]{000000} $0.3089$} & {\color[HTML]{000000} $0.4193$} \\
        \hline
        Likelihood & {\color[HTML]{000000} $0.38$} & {\color[HTML]{000000} $0.4982$} & {\color[HTML]{000000} $0.8788$} & {\color[HTML]{000000} $0.5528$} & {\color[HTML]{000000} $0.6433$} & {\color[HTML]{000000} $0.4396$} & {\color[HTML]{000000} $0.8077$} & {\color[HTML]{000000} $0.4596$} & {\color[HTML]{000000} $0.5690$} \\
        Entropy & {\color[HTML]{000000} $0.35$} & {\color[HTML]{000000} $0.6742$} & {\color[HTML]{000000} $0.3021$} & {\color[HTML]{000000} $0.5662$} & {\color[HTML]{000000} $0.5142$} & {\color[HTML]{000000} $0.6122$} & {\color[HTML]{000000} $0.2802$} & {\color[HTML]{000000} $0.5899$} & {\color[HTML]{000000} $0.4941$} \\
        LogRank & {\color[HTML]{000000} $0.36$} & {\color[HTML]{000000} $0.4711$} & {\color[HTML]{000000} $0.8496$} & {\color[HTML]{000000} $0.5597$} & {\color[HTML]{000000} $0.6268$} & {\color[HTML]{000000} $0.4002$} & {\color[HTML]{000000} $0.7694$} & {\color[HTML]{000000} $0.4472$} & {\color[HTML]{000000} $0.5389$} \\
        LRR & {\color[HTML]{000000} $0.41$} & {\color[HTML]{000000} $0.4016$} & {\color[HTML]{000000} $0.7203$} & {\color[HTML]{000000} $0.5629$} & {\color[HTML]{000000} $0.5616$} & {\color[HTML]{000000} $0.3095$} & {\color[HTML]{000000} $0.6214$} & {\color[HTML]{000000} $0.4710$} & {\color[HTML]{000000} $0.4673$} \\
        \note{DNA-GPT$\diamondsuit$} & {\color[HTML]{000000} $35.92$} & {\color[HTML]{000000} $0.5338$} & {\color[HTML]{000000} $0.8439$} & {\color[HTML]{000000} $0.3333$} & {\color[HTML]{000000} $0.5703$} & {\color[HTML]{000000} $0.4974$} & {\color[HTML]{000000} $0.7478$} & {\color[HTML]{000000} $0.3151$} & {\color[HTML]{000000} $0.5201$} \\   
        NPR$\diamondsuit$ & {\color[HTML]{000000} $111.99$} & {\color[HTML]{000000} $0.5659$} &{\color[HTML]{000000} $0.8786$} & {\color[HTML]{000000} $0.4246$} &{\color[HTML]{000000} $0.6230$} &{\color[HTML]{000000} $0.5065$} & {\color[HTML]{000000} $0.8444$} &{\color[HTML]{000000} $0.3740$} & {\color[HTML]{000000} $0.5750$}\\
        DetectGPT$\diamondsuit$ & {\color[HTML]{000000} $111.33$} & {\color[HTML]{000000} $0.6343$} &{\color[HTML]{000000} $0.8793$} & {\color[HTML]{000000} $0.5608$} &{\color[HTML]{000000} $0.6915$} &{\color[HTML]{000000} $0.6217$} & {\color[HTML]{000000} $0.8771$} &{\color[HTML]{000000} $0.5612$} & {\color[HTML]{000000} $0.6867$}\\
        Fast-Detect-GPT & {\color[HTML]{000000} $0.72$} & {\color[HTML]{000000} $0.7312$} & {\color[HTML]{000000} $0.9304$} & {\color[HTML]{000000} $0.7182$} & {\color[HTML]{000000} $0.7933$} & {\color[HTML]{000000} $0.6293$} & {\color[HTML]{000000} $0.8324$} & {\color[HTML]{000000} $0.6175$} & {\color[HTML]{000000} $0.6931$} \\
        \rowcolor[gray]{.9} 
        ImBD (Ours) & {\color[HTML]{000000} $0.72$} & \bf{\color[HTML]{000000} 
        \bm{$0.9849$}} & \bf{\color[HTML]{000000} \bm{$0.9871$}} & \bf{\color[HTML]{000000} \bm{$0.8626$}} & \bf{\color[HTML]{000000} \bm{$0.9449$}} & \bf{\color[HTML]{000000} \bm{$0.9486$}} & \bf{\color[HTML]{000000} \bm{$0.9468$}} & \bf{\color[HTML]{000000} \bm{$0.7743$}} & \bf{\color[HTML]{000000} \bm{$0.8899$}} \\
\hline

\hline

\hline

\hline
    \end{tabular}
        \caption{\textbf{Detection of \emph{GPT-3.5} and \emph{GPT-4o} polished text}. 
    Typically, the Neo-2.7B~\cite{black2021gpt} is used as the source for the scoring model. NPR and DetectGPT, on the other hand, utilize T5-3B~\cite{chen2019semantically} for generating perturbations, whereas Fast-DetectGPT employs GPT-J~\cite{wang2021gpt} as a surrogate model to generate samples. The $\diamondsuit$ symbol denotes methods that require multiple model invocations, leading to a substantial increase in computational load. Metric: AUROC. See Appendix B.5 for results on SQuAD.
}
    \label{tab:chatgpt_gpt4_results}
\end{table*}
\begin{table}[t]
\small
\centering
\setlength{\tabcolsep}{7pt}
\renewcommand\arraystretch{1.2}
\begin{tabular}{l|ccc|c}
\hline

\hline

\hline

\hline
        \multirow{1}{*}{\bf Method}
        & \bf XSum & \bf Writing & \bf PubMed & \bf Avg. \\
\hline

\hline
        \note{GPTZero} & \note{\color[HTML]{000000}$0.9542$} & \note{\color[HTML]{000000}$0.9711$} & \note{\bf {\color[HTML]{000000}\bm{$0.8800$}}} & \note{\color[HTML]{000000}$0.9351$} \\
        \rowcolor[gray]{.9} 
        ImBD (Ours) & \bf{\color[HTML]{000000} \bm{$0.9849$}} & \bf{\color[HTML]{000000} \bm{$0.9871$}} & {\color[HTML]{000000} {$0.8626$}} & \bf{\color[HTML]{000000} \bm{$0.9449$}}\\
\hline

\hline

\hline

\hline
    \end{tabular}
    \caption{\textbf{Compared with GPTZero on detecting \emph{GPT-3.5} polished text.} Metric: AUROC.
}
    \vspace{-0.1cm}
    \label{tab:gptzero}
\end{table}
\begin{table}[t]
\small
\centering
\setlength{\tabcolsep}{2pt}
\renewcommand\arraystretch{1.2}

    \centering\small
    \begin{tabular}{l|c@{\hspace{3.5pt}}c@{\hspace{3.5pt}}c@{\hspace{3.5pt}}c|c}
\hline

\hline

\hline

\hline
        \bf Method & \bf Qwen2 & \bf Llama-3 & \bf Mixtral & \bf Deepseek & \bf Avg. \\    
\hline

\hline
        Likelihood & {\color[HTML]{000000}$0.4121$} &  {\color[HTML]{000000} $0.6861$} & {\color[HTML]{000000}$0.5881$} & {\color[HTML]{000000}$0.6887$} & {\color[HTML]{000000}$0.5938$}  \\
        Entropy & {\color[HTML]{000000}$0.6819$} & {\color[HTML]{000000}$0.5546$} & {\color[HTML]{000000}$0.5741$} & {\color[HTML]{000000}$0.4923$} & {\color[HTML]{000000}$0.5757$}  \\
        LogRank & {\color[HTML]{000000}$0.3778$} & {\color[HTML]{000000}$0.6581$}& {\color[HTML]{000000}$0.5498$} & {\color[HTML]{000000}$0.6710$} & {\color[HTML]{000000}$0.5642$}  \\
        LRR & {\color[HTML]{000000}$0.3025$} & {\color[HTML]{000000}$0.5519$} & {\color[HTML]{000000}$0.4299$} & {\color[HTML]{000000}$0.6010$} & {\color[HTML]{000000}$0.4713$}  \\
        \note{DNA-GPT} & {\color[HTML]{000000} $0.5021$} & {\color[HTML]{000000}$0.6809$} & {\color[HTML]{000000}$0.6091$} & {\color[HTML]{000000}$0.7031$} & {\color[HTML]{000000}$0.6238$}  \\
        NPR  & {\color[HTML]{000000}$0.5388$} & {\color[HTML]{000000}$0.7186$} & {\color[HTML]{000000}$0.5988$}& {\color[HTML]{000000}$0.6551$} & {\color[HTML]{000000}$0.6278$}  \\
        DetectGPT  & {\color[HTML]{000000}$0.6193$} & {\color[HTML]{000000}$0.7706$} & {\color[HTML]{000000}$0.6826$} & {\color[HTML]{000000}$0.7160$} & {\color[HTML]{000000}$0.6971$}  \\
        Fast-DetectGPT & {\color[HTML]{000000}$0.7323$} & {\color[HTML]{000000}$0.8870$}& {\color[HTML]{000000}$0.8164$}  & {\color[HTML]{000000}$0.8687$} &  {\color[HTML]{000000}$0.8261$}  \\
        \rowcolor[gray]{.9} 
        ImBD (Ours) & \bf {\color[HTML]{000000}\bm{$0.9367$}}& \bf {\color[HTML]{000000}\bm{$0.9767$}} & \bf {\color[HTML]{000000}\bm{$0.9492$}} & \bf {\color[HTML]{000000}\bm{$0.9574$}} & \bf {\color[HTML]{000000}\bm{$0.9550$}}  \\

        \cline{1-6}
\hline

\hline

\hline

\hline
    \end{tabular}

    \caption{
    \textbf{
    Detection on \emph{open-source model} polished text.} AUROC scores are averaged across the XSum, SQuAD, and WritingPrompts datasets. Among them, Qwen2, Mixtral, and Deepseek are 7B models, while Llama-3 is an 8B model. Metric: AUROC. See Appendix B.3 for details.}
    \label{tab:main_results}
    \vspace{-0.5cm}
\end{table}
\begin{table}[t]
\small
\centering
\setlength{\tabcolsep}{3pt}
\renewcommand\arraystretch{1.2}

    \centering\small
    \begin{tabular}{l|c@{\hspace{3pt}}c@{\hspace{3pt}}c@{\hspace{3pt}}c@{\hspace{3pt}}|c}
\hline

\hline

\hline

\hline
        \multirow{2}{*}{\bf Method} & \multicolumn{4}{c}{\bf Tasks} \vrule & \multirow{2}{*}{\bf Avg.}\\
                & Rewrite & Expand & Polish & Generate & \\   
\hline

\hline
        Likelihood  & {\color[HTML]{000000}$0.4073$} & {\color[HTML]{000000} $0.4564$} & {\color[HTML]{000000}$0.6039$} & {\color[HTML]{000000}$0.8939$} & {\color[HTML]{000000}$0.5904$} \\
        Entropy & {\color[HTML]{000000}$0.5840$}  & {\color[HTML]{000000}$0.6629$} & {\color[HTML]{000000}$0.5431$} & {\color[HTML]{000000}$0.4129$}& {\color[HTML]{000000}$0.5507$}\\
        LogRank & {\color[HTML]{000000}$0.3868$} & {\color[HTML]{000000}$0.4273$} & {\color[HTML]{000000}$0.5864$} & {\color[HTML]{000000}$0.8925$}& {\color[HTML]{000000}$0.5732$}\\
        LRR & {\color[HTML]{000000}$0.3488$}  & {\color[HTML]{000000}$0.3581$} & {\color[HTML]{000000}$0.5183$}  & {\color[HTML]{000000}$0.8541$} & {\color[HTML]{000000}$0.5198$}\\
        \note{DNA-GPT} & {\color[HTML]{000000}$0.4101$}  & {\color[HTML]{000000}$0.4901$} & {\color[HTML]{000000}$0.5847$} & {\color[HTML]{000000}$0.8931$}& {\color[HTML]{000000}$0.5945$}\\
        NPR & {\color[HTML]{000000}$0.3606$}  & {\color[HTML]{000000}$0.5139$} & {\color[HTML]{000000}$0.5673$} & {\color[HTML]{000000}$0.8541$} & {\color[HTML]{000000}$0.5740$}\\
        DetectGPT   & {\color[HTML]{000000}$0.4060$} & {\color[HTML]{000000}$0.6000$} & {\color[HTML]{000000}$0.6615$} & {\color[HTML]{000000}$0.8985$} & {\color[HTML]{000000}$0.6415$}\\
        Fast-DetectGPT & {\color[HTML]{000000}$0.4499$}  & {\color[HTML]{000000}$0.7159$} & {\color[HTML]{000000}$0.7989$} & {\color[HTML]{000000}$0.9706$} & {\color[HTML]{000000}$0.7338$}\\
        \rowcolor[gray]{.9} 
        ImBD (Ours) & \bf {\color[HTML]{000000}\bm{$0.8739$}} & \bf {\color[HTML]{000000}\bm{$0.9758$}} & \bf {\color[HTML]{000000}\bm{$0.9707$}} & \bf {\color[HTML]{000000}\bm{$0.9996$}} & \bf {\color[HTML]{000000}\bm{$0.9550$}}\\
\hline

\hline

\hline

\hline
    \end{tabular}
    \caption{
\textbf{Performance on diverse tasks.} 
We evaluated the detection performance, measured by average AUROC, of text revised by leading LLMs (Qwen2-7B, Llama-3-8B, Mixtral-7B, Deepseek-7B, GPT-3.5, and GPT-4o) on the XSum dataset.
    }
    \vspace{-0.5cm}
    \label{tab:domain_generlization}
\end{table}
\begin{table}[t]
\small
\centering
\setlength{\tabcolsep}{1.2pt}
\renewcommand\arraystretch{1.2}
    \centering\small
    \begin{tabular}{l|c@{\hspace{1.2pt}}c@{\hspace{1pt}}c@{\hspace{1.2pt}}c|c@{\hspace{1.2pt}}c@{\hspace{1.2pt}}c@{\hspace{1.2pt}}c}
\hline

\hline

\hline

\hline
        \multirow{2}{*}{\bf Strategy}& \multicolumn{3}{c}{\bf GPT-3.5} &  \multirow{2}{*}{ \bf Avg.}&\multicolumn{3}{c}{\bf GPT-4o}&  \multirow{2}{*}{ \bf Avg.}  \\
         & XSum & Writing & Pub. & & XSum & Writing & Pub. &  \\ 
\hline
        w/o imitate & {\color[HTML]{000000}$0.73$} &  {\color[HTML]{000000}$0.93$} & {\color[HTML]{000000}$0.72$} & {\color[HTML]{000000}$0.79$} & {\color[HTML]{000000}$0.63$} & {\color[HTML]{000000}$0.83$} & {\color[HTML]{000000}$0.62$} & {\color[HTML]{000000}$0.69$}  \\
        SFT & {\color[HTML]{000000}0.56} &  {\color[HTML]{000000}$0.70$} & {\color[HTML]{000000}$0.70$} & {\color[HTML]{000000}$0.65$} & {\color[HTML]{000000}$0.60$} & {\color[HTML]{000000}$0.74$} & {\color[HTML]{000000}$0.66$} & {\color[HTML]{000000}$0.67$}  \\
        SFT* &{\color[HTML]{000000}$0.59$} & {\color[HTML]{000000}$0.70$}& {\color[HTML]{000000}$0.66$} & {\color[HTML]{000000}$0.65$} & {\color[HTML]{000000}$0.61$} & {\color[HTML]{000000}$0.73$} & {\color[HTML]{000000}$0.60$} & {\color[HTML]{000000}$0.65$} \\
        RLHF &  {\color[HTML]{000000}$0.70$} & {\color[HTML]{000000}$0.92$}& {\color[HTML]{000000}$0.78$} & {\color[HTML]{000000}$0.80$} & {\color[HTML]{000000}$0.54$} & {\color[HTML]{000000}$0.81$} & {\color[HTML]{000000}$0.64$} & {\color[HTML]{000000}$0.66$}\\
        ORPO &  {\color[HTML]{000000}0.79} & {\color[HTML]{000000}$0.97$}& {\color[HTML]{000000}$0.81$} & {\color[HTML]{000000}$0.86$} & {\color[HTML]{000000}$0.60$} & {\color[HTML]{000000}$0.87$} & {\color[HTML]{000000}$0.66$} & {\color[HTML]{000000}$0.71$}\\
         \rowcolor[gray]{.9} 
        ImBD (Ours) &  \bf {\color[HTML]{000000} 0.99} & \bf {\color[HTML]{000000} \bm{$0.99$}} & \bf {\color[HTML]{000000} \bm{$0.86$}} & \bf {\color[HTML]{000000} \bm{$0.95$}}& \bf{\color[HTML]{000000} \bm{$0.95$}} & \bf{\color[HTML]{000000} \bm{$0.95$}} & \bf{\color[HTML]{000000} \bm{$0.77$}} & \bf{\color[HTML]{000000} \bm{$0.89$}} \\
        \cline{1-4}
\hline

\hline

\hline

\hline
    \end{tabular}
  \caption{\textbf{Ablation on preference optimization.} Comparative performance of SPO, supervised fine-tuning (SFT), RLHF, and ORPO strategies across datasets. Training dataset size: $1,000$ samples. ``*" denotes trained on 3x samples. ``Pub." denotes ``PubMed". Metric: AUROC. Task: \texttt{Polish}.}
    \label{tab:SPO VS. LLM finetune}
\end{table}
\subsection{Machine revision dataset}
\paragraph{Data sources} 
The human-written texts included in the training dataset were crawled from the internet before 2019. The texts are then polished by GPT-3.5.\footnote{All mentions of GPT-3.5 and GPT-4o in this paper refer to \texttt{gpt-3.5-turbo-0125} and \texttt{gpt-4o-2024-05-13}, respectively.} We use $500$ pairs of samples for training. The composition of the dataset is $57.3$\% papers, $14.2$\% blogs, $4.0$\% letters and emails, and $2.1$\% homework. See Appendix A.1 and A.2 for training and dataset collection details, respectively.

For the test data, we follow ~\citet{bao2023fast,mitchell2023detectgpt}, use paragraphs from diverse domains as human-written texts, including \emph{XSum}~\citep{narayan2018don} for news articles, \emph{SQuAD}~\citep{fan2018hierarchical} for Wikipedia contexts, \emph{WritingPrompts}~\citep{fan2018hierarchical} (Abbreviated as ``Writing") for story writing, and \emph{PubMedQA}~\citep{jin2019pubmedqa} for biomedical research question answering. Then, we use the pipeline detailed in the following paragraph to generate correspondent machine-revised text.

\paragraph{Dataset process}
\label{sec:dataset_process}
We design a cohesive two-stage pipeline to revise human-written text. Detailed examples of the generated instructions are in Appendix A.3.
\begin{itemize}
    \item \textbf{Revision instruction generation}: For each task, instructions are constructed with varying tones and lengths using GPT-3.5. The tone is randomly selected from a set of $10$ predefined options, while the instruction length is chosen from the set of
    $\{15, 30, 50\}$ words. The intuition behind choosing different tones and lengths is to simulate different human behaviors.
    \item \textbf{Paragraph revision}: The generated instruction and the human-written text are then prompted into the LLM to produce the final machine-revised text.
\end{itemize}
\paragraph{Target LLMs for revision}
We experiment with four open-source models: Qwen2-7B~\citep{yang2024qwen2technicalreport}, Llama-3-8B~\citep{llama38b}, Mixtral-7B~\citep{jiang2024mixtralexperts}, and Deepseek-7B~\citep{deepseekai2024deepseekllmscalingopensource}, as well as two proprietary models, GPT-3.5~\citep{chatgpt} and GPT-4o~\citep{openai2024gpt4}. Our choice covers a broad spectrum of user preferences. 

\paragraph{Machine revision tasks}
We evaluate the performance of the detector on three tasks: \texttt{rewrite}, \texttt{expand}, and \texttt{polish}. 
\begin{itemize}
    \item \textbf{\texttt{Rewrite}}: The LLM is asked to rewrite the given text while preserving all details.
    \item \textbf{\texttt{Expand}}: The LLM is asked to expand the original text given a style parameter randomly chosen from a set of $10$ options such as formal, literary, \textit{etc}. 
    \item \textbf{\texttt{Polish}}: The LLM is asked to polish/adjust the text based on a randomly picked style. 
\end{itemize}
Furthermore, we test our method on the \texttt{generate} task used in the common evaluation of machine-generated text detectors, which does not fall under the category of machine-revised text detection. To produce machine-generated text for \texttt{generate} task, the LLM is prompted with the first $30$ tokens of the human written text, following the design in DetectGPT~\citep{mitchell2023detectgpt} and Fast-DetectGPT~\citep{bao2023fast}. The details on generating those instructions can be found in Appendix A.3.

\subsection{Baselines}
We compare our method with two lines of method: training-based models, and logit-based models.
Following~\citet{bao2023fast}, we use AUROC as a metric to evaluate detection accuracy.
\begin{itemize}
    \item \textbf{Training-based models} include RoBERTa-base~\citep{liu2019roberta} and RoBERTa-large~\cite{liu2019roberta}, which is trained on substantial datasets up to $160$GB of text data, as well as the commercial detector GPTZero~\citep{tian2023gptzero}, which is trained on massive datasets.
    \item \textbf{Logit-based models} include 
    Likelihood~\citep{ippolito2020automatic} (mean log probabilities), LogRank~\citep{solaiman2019release} (average log of ranks in descending order by probabilities), Entropy~\citep{gehrmann2019gltr} (mean token entropy of the predictive distribution), LRR~\citep{su2023detectllm} (an amalgamation of log probability and log-rank), NPR~\citep{su2023detectllm} (normalized perturbed log-Rank)  \note{and DNA-GPT~\citep{yang2023dna}} (divergent N-Gram Analysis), DetectGPT~\citep{mitchell2023detectgpt}, and its advanced variant, Fast-DetectGPT~\citep{bao2023fast}.
\end{itemize}
Note that Fast-DetectGPT~\citep{bao2023fast}, the current state-of-the-art approach, also serves as a baseline method that does not involve machine-style imitation.
\subsection{Main results}
\paragraph{Detection performance for GPT series} We evaluate our method using passages polished by GPT-3.5 and GPT-4o across different domains. As shown in Table~\ref{tab:chatgpt_gpt4_results}, our method outperforms Fast-DetectGPT by $15.16$\% and $19.68$\% in detecting GPT-3.5 and GPT-4 outputs, respectively, on the \texttt{polish} task. See Appendix B.1 for ROC curves of detecting GPT-3.5 and GPT-4 texts.
Furthermore, compared to the supervised detectors RoBERTa-large, our method shows an improvement of $32.91$\%/$47.06$\% on detecting GPT-3.5 and GPT-4, respectively. Additionally, as shown in Table~\ref{tab:gptzero}, our method surpasses GPTZero by $0.98$\%. This indicates that our method is highly efficient in training, achieving superior performance with a small amount of data compared to models trained on much larger datasets. 
To demonstrate task generalization, we compared performance on the \texttt{rewrite} task, where our method outperformed Fast-DetectGPT by $36.96$\% and $24.29$\% in detecting GPT-3.5 and GPT-4o outputs, respectively See Appendix B.2 for detail results on each dataset, Appendix B.3 for the performance testing on \texttt{rewrite} task, and Appendix B.5 for results on SQuAD dataset. See Appendix B.6 for experimental results in multilingualism.

\paragraph{Detection performance on open-source models} 
The performance on \texttt{polish} task by open-source models is shown in Table~\ref{tab:main_results}. ImBD achieves the highest average AUROC, outperforming DetectGPT by $25.79$\%. 
The detailed results are included in Appendix B.2.

\paragraph{Robustness in machine revision and generation}
As shown in Table~\ref{tab:domain_generlization}, our method outperforms the state-of-art Fast-DetectGPT by $22.12$\% on average across all four tasks. The results showcase the robustness of our approach across various tasks and user instructions. See Appendix B.4 for detailed results.

\paragraph{Inference time and training efficiency}

Our model is trained for $2$ epochs with a learning rate set to $0.0001$ and $\beta$ set to $0.05$. Each epoch requires approximately $110$ seconds on an L20 (48G) GPU, leading to a total training time of $220.57$ seconds. As shown in Table~\ref{tab:chatgpt_gpt4_results}
, our method achieves a competitive inference time of $0.72$ seconds per $1000$ words, matching that of Fast-DetectGPT ($154.62\times$ speed-up compared to DetectGPT), but with better performance.
\begin{figure}[t]
    \centering
    \includegraphics[width=1.0\linewidth]{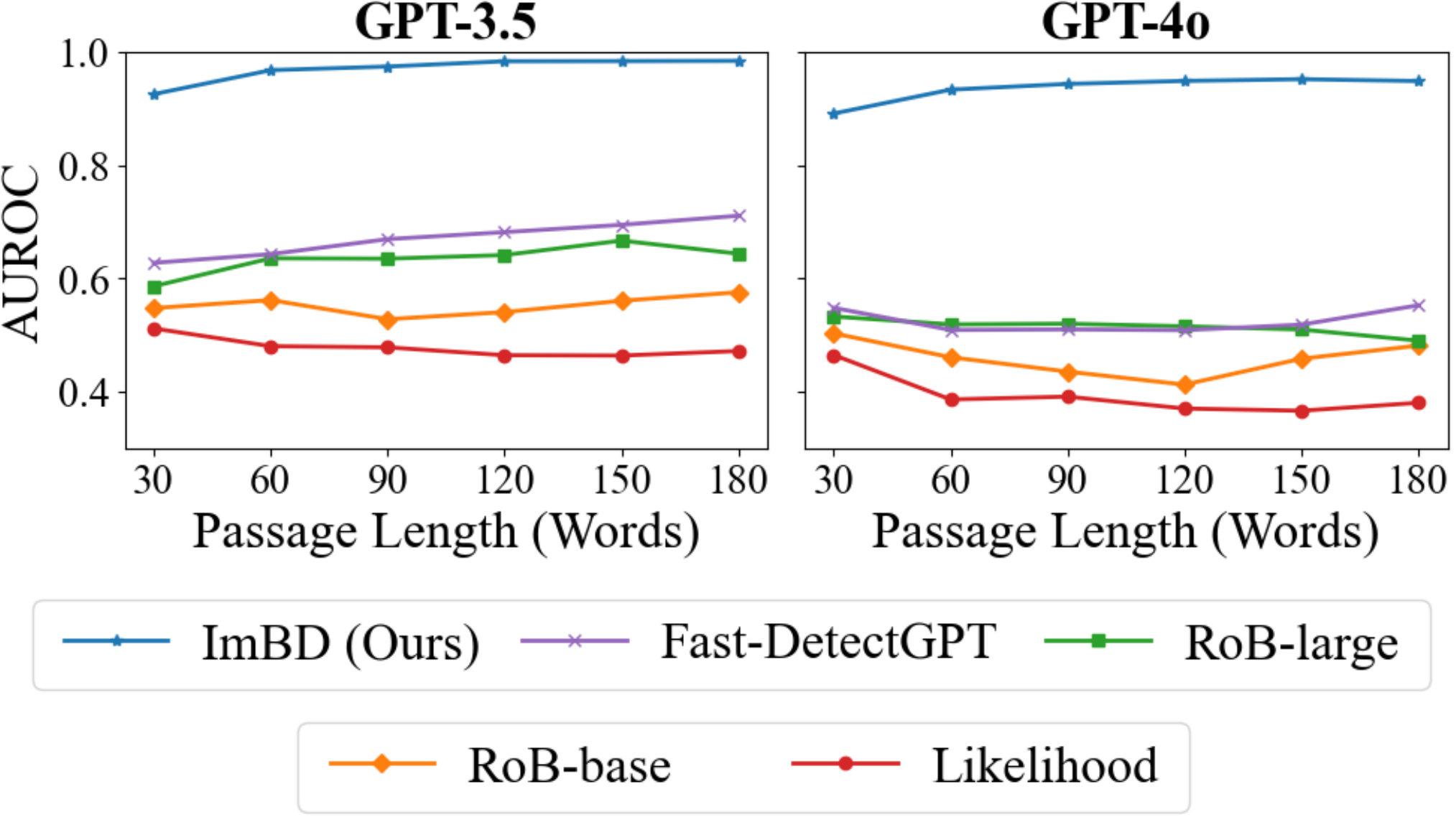}
    \caption{Evaluations of detection accuracy for XSum polished texts trimmed to the specified word count.}
    \vspace{-0.5cm}
    \label{fig:AUROC-PL}
\end{figure}

\subsection{Ablation study}
\paragraph{Ablation on machine-style imitation}
As shown in Table~\ref{tab:SPO VS. LLM finetune}, using fast-DetectGPT as the baseline without imitation, our method improves detection accuracy by 16\% and 20\% on GPT-3.5 and GPT-4o machine-revised texts, respectively.
\paragraph{Ablation on preference optimization}
To demonstrate the difference between different optimization methods on ImBD, we compare the performance of SPO against other alignment approaches on \texttt{polish} task.
As shown in Table~\ref{tab:SPO VS. LLM finetune}, ImBD outperformed the SFT variant by $30$\% on GPT-3.5 and $24$\% on GPT-4o, even when the SFT variant uses $3$x training data. Additionally, ImBD exceeds RLHF and ORPO significantly.

\paragraph{Ablation on text length}
As shown in Figure~\ref{fig:AUROC-PL}, our method demonstrates strong performance across passages of varying lengths compared to other methods, with accuracy improving as passage length increases.

\section{Related work}
\subsection{Machine-Generated Text Detection}
\paragraph{Datasets}
Researchers developed various evaluation benchmarks for machine-generated text detection. \citet{bao2023fast} and \citet{mitchell2023detectgpt} used the initial 30 tokens from human-written texts across different domains as prompts to generate pure machine-generated text via LLMs. Following this approach, \citet{guo2023closechatgpthumanexperts} employed QA datasets as human samples and generated pure machine-generated text using ChatGPT. Building upon the QA framework, researchers~\citep{mitchell2023detectgpt,su2023detectllm,NEURIPS2023_30e15e59,he2024mgtbench, Wang2024m4} collected texts generated by mainstream LLMs. \citet{Verma2023ghostbuster} focused on creative writing tasks, providing only writing prompts or headlines to generate text with LLMs. However, a significant portion of contemporary machine-generated content involves human input~\cite{Zhang2024mixtext}. For instance, MixSet~\citep{Zhang2024mixtext} examined scenarios where human revisions are applied to machine-generated text. In contrast, our study focuses on the reverse: human-written text revised by LLMs. This practice, where people use AI to enhance, edit, or expand their writing, is increasingly common and accepted in various contexts but remains largely prohibited in academic settings. We specifically address detecting this form of human-machine collaborative text.
\paragraph{Methods}
Previous methods for machine-generated text detection generally fall into two categories: training-based methods and logit-based metric approaches. While training-based methods~\citep{guo2023closechatgpthumanexperts,chen2023gpt,NEURIPS2023_30e15e59} achieved excellent performance due to large-scale data and high-cost training, they tended to overfit and were less effective in detecting the machine-revised text.

Existing logit-based approaches, such as Log-Likelihood~\citep{solaiman2019release}, Entropy~\citep{solaiman2019release}, Rank~\citep{gehrmann2019gltr}, and Log-Rank~\citep{mitchell2023detectgpt}, relied on statistical analysis to evaluate information beyond the token level. GLTR~\citep{gehrmann2019gltr} combined a set of metric-based methods to assist human identification. DetectGPT~\citep{mitchell2023detectgpt} built on the observation that machine-generated texts occupy regions with steep negative log probability curvature, using this probability curvature to detect whether text originates from LLMs. This concept was further developed and improved in subsequent studies~\cite{su2023detectllm, Mireshghallah2023slm, bao2023fast}. \citet{zeng2024improving} proposed adapting scoring models through fine-tuning to handle the latest black-box models.

While previous approaches generally relied on overall text features for classification, we propose isolating stylistic features as the basis, enabling more precise detection of subtle differences.

\subsection{Preference Optimization}
Direct Preference Optimization~\citep{rafailov2023directpreferenceoptimizationlanguage} can efficiently learn and align preferences from a pair of sampled texts. Related offline algorithms~\citep{yuan2024rrhf,ethayarajh2024ktomodelalignmentprospect,hong2024orpo,park2024disentangling} were typically also employed to align LLMs with human preferences, primarily for text-generation tasks. However, our study is the first to apply preference optimization methods to align with a distinct AI style (rather than aligning with human preferences) and to use this approach for classification in the context of machine-revised text detection.

\section{Conclusion}
In this work, we have presented the ``\textit{\textbf{Imitate Before Detect}}" paradigm to detect machine-revised text by learning to imitate the writing style of LLMs.
Specifically,  we have proposed style preference optimization for aligning the detector with machine writing styles and leveraged style-conditional probability curvature to quantify log probability differences for effective detection.
We have conducted extensive evaluations across six leading LLMs, three text domains, and three revision techniques, demonstrating significant improvements in detection accuracy compared to existing state-of-the-art methods.

\section*{Acknowledge}
We express our gratitude to Fenz.AI for their research funding and to Zhenyu Ding, Yuanhe Chang, and Longzhi Bing from MercallureAI for providing the computational platform. We also appreciate the significant contributions to data collection by Fulong Yang, alongside the research and deployment support from Yue Wang and Yifei Ke.
\bibliography{cited}

\appendix
\maketitlesupplementary
\section{Implement details}
\subsection{Training details}
We fine-tune the \texttt{gpt-neo-2.7B} model from \texttt{EleutherAI}, using a learning rate of $0.0001$ and a beta coefficient of $0.05$. The fine-tuning process is conducted over $2$ epochs ($1,000$ samples each epoch) with a fixed random seed of $42$ to ensure reproducibility.
For parameter-efficient training, we utilize a Lora configuration with a rank of $8$, a Lora alpha of $32$, and a dropout rate of $0.1$, specifically tailored for causal language modeling tasks. Note that, the learning rate is the best choice from $\{ 0.1, 0.01, 0.001, 0.0001, 0.00001\}$.
All experiments are conducted on an Ubuntu 20.04 platform using a single L20 (48GB) GPU, with Python 3.8, PyTorch 1.10.0, Transformers 4.28.1, and Datasets 2.12.0.

\subsection{Dataset collection details}
To train a detector with strong generalization capabilities, we do not use existing domain-specific text datasets. Instead, we randomly collect $500$ paragraphs from the internet, all published before 2019. Each paragraph is approximately $300$ words long and covers one of seven topics: academic papers, assignments, blogs, letters, literary works, news articles, and others. These texts represent human-authored content from before 2019. We then processed these human-written texts through a \texttt{polish} data generation pipeline by GPT-3.5-turbo, resulting in $500$ pairs of human and machine-revised texts. All experimental data in the main paper are based on these pairs. Note that all data were manually collected, with collectors compensated at a rate of \$ $60$ per hour. We ensured that no copyrighted texts were used, and the model trained on this data is intended solely for academic discussion, with no commercial use planned.

\subsection{Machine-revised text generation pipeline}\label{tasks_prompt}
We design a two-step generation pipeline for machine-revised text data generation. First, the pipeline generates a user instruction with GPT-3.5-turbo, and then we combine human-written text and the generated user instruction, to generate the machine-revised texts.

\paragraph{Data generation pipeline for \texttt{rewrite} task}
Prompt template for \texttt{rewrite task}:
\begin{quote}
\begin{scriptsize}
\begin{verbatim}
"You are a professional rewriting expert and you can 
help paraphrasing this paragraph in English without 
missing the original details.  Please keep the length 
of the rewritten text similar to the original text. 
<original>"
\end{verbatim}
\end{scriptsize}
\end{quote}

The $<$original$>$ could be like: 
\begin{quote}
\begin{scriptsize}\begin{verbatim}
"Chief executive Bimlendra Jha praised the `significant 
effort' to turn things around - after Port Talbot was 
said to be losing Â£1m a day earlier this year. Tata
looked to sell its UK business but paused the process 
in July. Losses at Port Talbot have been reduced by 
a turnaround plan and better market conditions such 
as rising steel prices and a drop in the pound's value. 
Mr Jha also said he expects the UK government to keep 
a promise of helping to solve the company's pension 
problems. However, he warned: `We must remember that 
whether you drown one foot under the water or ten feet 
under the water, you still drown.' A Â£485m pension 
deficit was given as a reason for some potential buyers 
being put off. The UK government under the previous 
prime minister, David Cameron, launched a consultation 
that would have brought in legal changes to the scheme. 
It would have reduced its}".
\end{verbatim}\end{scriptsize}
\end{quote}
The $<$original$>$ serves as the original human-written text from the dataset;

\paragraph{Data generation pipeline for \texttt{polish} task}
Prompt template and candidate parameters for \texttt{polish} task:
\begin{quote}
\begin{scriptsize}\begin{verbatim}
word_lens = [15,30,50]
styles = ["formal", "oral", "academic", "literary", 
"critical", "narrative", "descriptive", "lyric", 
"objective", "subjective"]
<word_len> = random.choice(word_lens)
<style> = random.choice(styles)
"Write a prompt in <word_len> words that says  you want 
gpt's help in polishing a paragraph in a <style> style,
this prompt can only be <word_len> words or less."
\end{verbatim}\end{scriptsize}
\end{quote}

In the task of text polish, we initially define three types of ``word\_len" ($15$ words, $30$ words, and $50$ words) for the prompt and a range of ``styles" (including formal, oral, academic, literary, critical, narrative, descriptive, lyric, objective, and subjective) to designate distinct textual styles. Subsequently, we randomly ascertain the requisite word count and stylistic choice from these parameters and craft a prompt based on these criteria. This prompt is then utilized to generate a secondary prompt intended to the polish task.
\begin{quote}
\begin{scriptsize}\begin{verbatim}
`<prompt>\n<original>'
\end{verbatim}\end{scriptsize}
\end{quote}

\paragraph{Data generation pipeline for \texttt{expand} task}
The $<$prompt$>$ is the output generated in the previous step. The $<$prompt$>$ could be like: ``Need help refining paragraph with vivid descriptions for a more polished piece. Assist me, please." The prompt template and candidate parameters for \texttt{expand} task are the following:
\begin{quote}
\begin{scriptsize}
\begin{verbatim}
styles = ["formal", "oral", "academic", "literary",
"critical", "narrative", "descriptive", "lyric",
"objective", "subjective", "original"]
<style> = random.choice(styles)
`Expand but not extend the paragraph in a <style> style.
\n<original> `The expanded paragraph:'
\end{verbatim}
\end{scriptsize}
\end{quote}

\paragraph{Data generation pipeline for \texttt{generate} task}
Prompt template for \texttt{generate} task is following:
\begin{quote}
\begin{scriptsize}\begin{verbatim}
"You are a News writer. Please write an article 
with about 150 words starting exactly with: <prefix>"
\end{verbatim}\end{scriptsize}
\end{quote}

The $<$prefix$>$ is the first 30 tokens of the human-written sentence. The $<$prefix$>$ could be like: 
\begin{quote}
\begin{scriptsize}\begin{verbatim}
"Chief executive Bimlendra Jha praised the `significant 
effort' to turn things around - after Port Talbot was 
said to be losing Â"
\end{verbatim}\end{scriptsize}
\end{quote}

\section{Additional results}
\label{additional_results}
\subsection{ROC Curve Analysis on XSum Dataset}
\begin{figure}[t]
    \centering
    \includegraphics[width=1.0\linewidth]{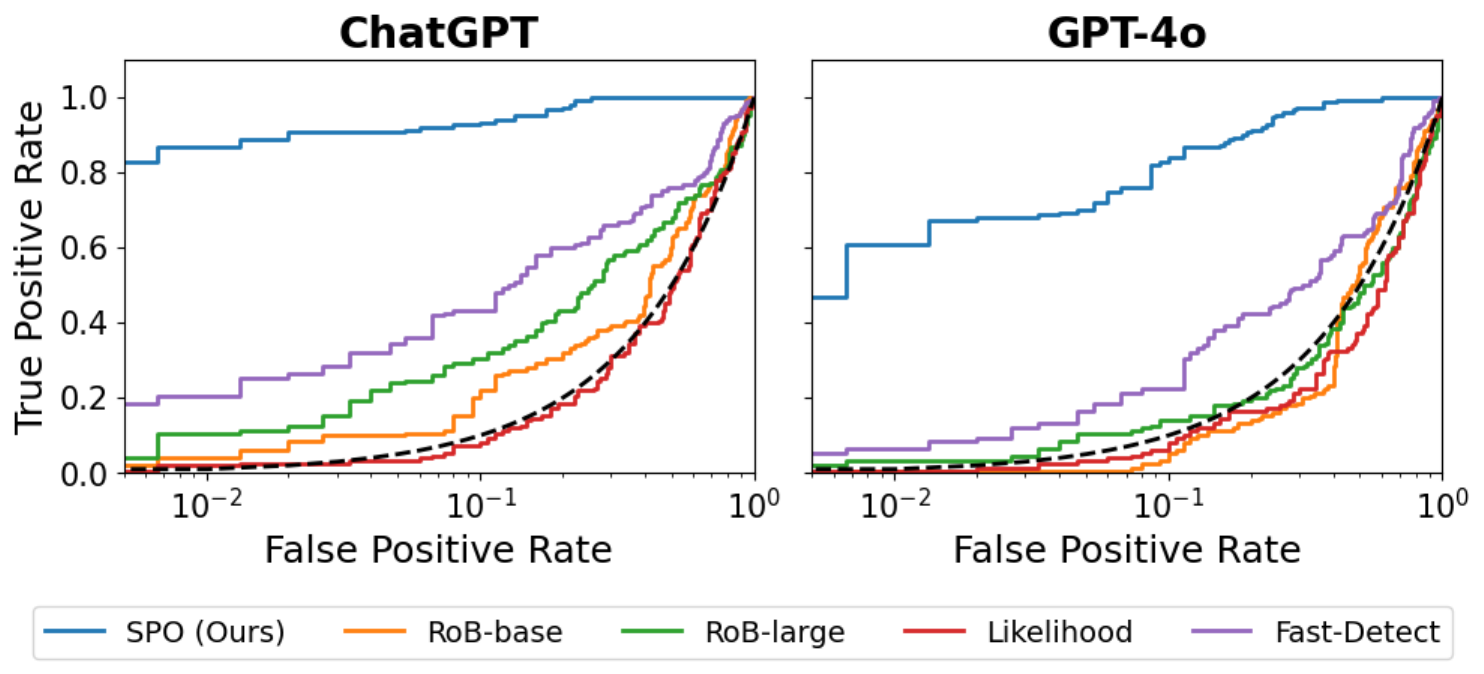}
    \caption{ROC curves in log scale evaluated on \texttt{polish} task of XSum dataset, where the dash lines denote the random classifier. ``Fast-Det." denotes ``Fast-DetectGPT".1}
    \label{fig:roc}
\end{figure}

As shown in Figure~\ref{fig:roc}, the ROC curves in Figure 1 demonstrate the performance of various detection methods on the XSum dataset, evaluated on both ChatGPT and GPT-4o, where our method (SPO) consistently outperforms others across different false positive rates, with the dashed lines indicating the random classifier's performance.

\subsection{Detail results of machine-polished text detection}
The results in Table~\ref{tab:main_result_details_polish} demonstrate the outstanding performance of our method (ImBD) in detecting text generated by open-source models. ImBD achieved the highest AUROC scores across all datasets, including XSUM, SQuAD, and WritingPrompts, and consistently outperformed other methods regardless of the target model (Qwen2-7B, Llama-3-8B, Mistral-7B, or Deepseek-7B).
\subsection{Detail results of machine-rewitten text detection}\label{open_source}
Table~\ref{tab:chatgpt_gpt4_rewritten_results} and Table~\ref{tab:main_result_details_rewritten} compare various methods in detecting rewritten texts generated by GPT-3.5, GPT-4
and \texttt{gpt-4o-2024-05-13}, and four popular open-source LLMs. Our method (ImBD) performed exceptionally well across all tasks, achieving the highest accuracy.

\subsection{Detail results on diverse machine-revision tasks and target LLMs}
The results in Table~\ref{tab:domain_generlization_details} further demonstrate the exceptional performance of our method across pure machine text generation and various text revision tasks, including \texttt{rewrite}, \texttt{polish}, and \texttt{expand}. The ImBD consistently outperformed other methods across all combinations of models and tasks, proving its broad applicability and robust detection capability in diverse text revision tasks. Notably, ImBD was able to detect outputs from various target LLMs effectively, highlighting its strong generalization capabilities not only for individual tasks but also when addressing complex and varied text generation challenges across different models.
\subsection{Additional results on detecting GPT-3.5}
\begin{figure*}[t]
    \centering
    \includegraphics[width=1.0\linewidth]{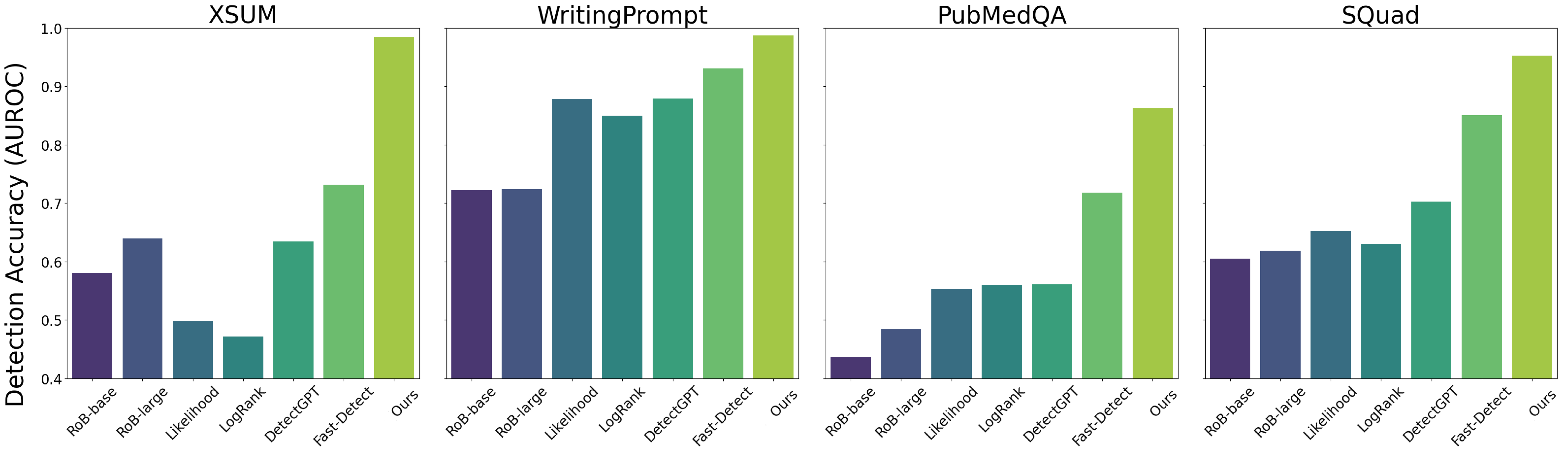}
    \caption{\textbf{Additional performance comparison on detecting machine-polished text.} Target LLM: GPT-3.5.}
    \label{fig:AUROC}
\end{figure*}

We present additional results comparing the performance of various methods in detecting machine-polished text across four datasets, including XSUM, WritingPrompt, PubMedQA, and SQuAD. Notably, the inclusion of SQuAD (news articles) allows for a more comprehensive evaluation. As shown in Figure~\ref{fig:AUROC}, our method consistently outperforms previous approaches across all datasets, demonstrating its superior detection capability in identifying GPT-3.5 polished text. This further underscores the robustness and generalizability of our approach in different contexts.

\subsection{Robustness on diverse linguistic properties}
We conducted some additional experiments on machine-revised text in Spanish~\cite{maria}, Portuguese~\cite{sharegpt-portuguese}, and Chinese~\cite{wang2024humanchineseppo}. The experimental setup is the same as in the main paper, training on $500$ pairs of texts and testing on $100$ pairs of texts. All data from public datasets and internet blogs for diversity. The results (shown in table~\ref{tab:diverse_languages}) indicate that ImBD consistently outperforms Fast-DetectGPT across all three languages.
\begin{table}[t]
\small
\centering
\setlength{\tabcolsep}{13pt}
\renewcommand\arraystretch{1.2}

    \centering\small
    \begin{tabular}{l|c@{\hspace{12.5pt}}c@{\hspace{12.5pt}}c}
\hline

\hline

\hline

\hline
        \bf Method & \bf Spanish & \bf Portuguese & \bf Chinese\\    
\hline

\hline
        Likelihood & {\color[HTML]{000000}$0.6423$} &  {\color[HTML]{000000} $0.5580$} & {\color[HTML]{000000}$0.8129$}\\
        Entropy & {\color[HTML]{000000}$0.4209$} & {\color[HTML]{000000}$0.4918$}& {\color[HTML]{000000}$0.2381$}\\
        LogRank & {\color[HTML]{000000}$0.6212$} & {\color[HTML]{000000}$0.5414$} & {\color[HTML]{000000}$0.8118$}\\
        LRR & {\color[HTML]{000000}$0.5110$} & {\color[HTML]{000000}$0.7354$} & {\color[HTML]{000000}0.4720}\\
        \note{DNA-GPT} & {\color[HTML]{000000} $0.5350$} & {\color[HTML]{000000}$0.4313$} & {\color[HTML]{000000}-}\\
        NPR  & {\color[HTML]{000000}$0.6632$} & {\color[HTML]{000000}$0.6452$} & {\color[HTML]{000000}$0.5001$}\\
        DetectGPT  & {\color[HTML]{000000}$0.3820$} & {\color[HTML]{000000}$0.3750$} & {\color[HTML]{000000}$0.5001$} \\
        Fast-DetectGPT & {\color[HTML]{000000}$0.6627$} & {\color[HTML]{000000}$0.5445$}& {\color[HTML]{000000}$0.8060$}\\
        \rowcolor[gray]{.9} 
        ImBD (Ours) & \bf {\color[HTML]{000000}\bm{$0.8487$}}& \bf {\color[HTML]{000000}\bm{$0.8214$}} & \bf {\color[HTML]{000000}\bm{$0.8792$}} \\
\hline

\hline

\hline

\hline
    \end{tabular}

    \caption{
    \textbf{
     Performance on diverse languages.} We evaluated the detection performance of text polished by GPT-3.5 across Spanish, Portuguese, and Chinese from public datasets and internet blogs. Metric: AUROC. ``-'' means the model fail on this language.}
    \label{tab:diverse_languages}
    \vspace{-0.5cm}
\end{table}
\begin{table*}[t]
    \centering\small
    \setlength{\tabcolsep}{7pt}
    \begin{tabular}{l@{\hspace{10.5pt}}l@{\hspace{10.5pt}}c@{\hspace{10.5pt}}c@{\hspace{10.5pt}}c@{\hspace{10.5pt}}c@{\hspace{10.5pt}}c}
\hline

\hline

\hline

\hline
        \multirow{2}{*}{\bf Dataset} & \multirow{2}{*}{\bf Method} & \multicolumn{4}{c}{\bf Source Model} & \multirow{2}{*}{\bf Avg.}\\
         &   &  Qwen2-7B &  Llama-3-8B &  Mixtral-7B &  Deepseek-7B & \\    
        \midrule
        \multirow{10}{*}{XSum} 
        & Likelihood~\citep{ippolito2020automatic} & {\color[HTML]{000000}$0.2520$} & {\color[HTML]{000000}$0.5695$} & {\color[HTML]{000000}$0.4353$} & {\color[HTML]{000000}$0.5438$} & {\color[HTML]{000000}$0.4502$}  \\
        & Entropy~\citep{gehrmann2019gltr} & \underline{\color[HTML]{000000}$0.7623$ } & {\color[HTML]{000000}$0.6348$} & {\color[HTML]{000000}$0.6539$} & {\color[HTML]{000000}$0.6402$} & {\color[HTML]{000000}$0.6728$}  \\
        & LogRank~\citep{solaiman2019release} & {\color[HTML]{000000}$0.2246$} & {\color[HTML]{000000}$0.5412$} & {\color[HTML]{000000}$0.3980$} & {\color[HTML]{000000}$0.5288$} & {\color[HTML]{000000}$0.4232$}  \\
        & LRR~\citep{su2023detectllm} & {\color[HTML]{000000}$0.1875$} & {\color[HTML]{000000}$0.4530$} & {\color[HTML]{000000}$0.3112$} & {\color[HTML]{000000}$0.4859$} & {\color[HTML]{000000}$0.3594$}  \\
        & \note{DNA-GPT~\citep{yang2023dna} $\diamondsuit$} & {\color[HTML]{000000}$0.3352$} & {\color[HTML]{000000}$0.5599$} & {\color[HTML]{000000}$0.4555$} & {\color[HTML]{000000}$0.5586$} & {\color[HTML]{000000}$0.4773$}  \\
        & NPR~\citep{su2023detectllm} $\diamondsuit$ & {\color[HTML]{000000}$0.3896$} & {\color[HTML]{000000}$0.6144$} & {\color[HTML]{000000}$0.4594$} & {\color[HTML]{000000}$0.5476$} & {\color[HTML]{000000}$0.5028$}  \\
        & DetectGPT~\citep{mitchell2023detectgpt} $\diamondsuit$ & {\color[HTML]{000000}$0.4885$} & {\color[HTML]{000000}$0.6904$} & {\color[HTML]{000000}$0.5480$} & {\color[HTML]{000000}$0.6172$} & {\color[HTML]{000000}$0.5860$}  \\
        & Fast-DetectGPT~\citep{bao2023fast} & {\color[HTML]{000000}$0.5945$} & \underline{\color[HTML]{000000}$0.8192$ } & \underline{\color[HTML]{000000}$0.7034$ } & \underline{\color[HTML]{000000}$0.8177$ } & \underline{\color[HTML]{000000}$0.7337$ }  \\
        \rowcolor[gray]{.9}
        & ImBD (Ours) & \bf {\color[HTML]{000000} \bm{$0.9589$}} & \bf  {\color[HTML]{000000}\bm{$0.9884$}} & \bf {\color[HTML]{000000} \bm{$0.9671$}} & \bf {\color[HTML]{000000}\bm{$0.9764$}} & \bf {\color[HTML]{000000}\bm{$0.9727$}}  \\
        \rowcolor[gray]{.9}
        & \it (Diff) & \textit {\color[HTML]{000000}$\mathit{0.1966}$} & \it {\color[HTML]{000000}$\mathit{0.1692}$} & \it {\color[HTML]{000000}$\mathit{0.2637}$} & \it {\color[HTML]{000000}$\mathit{0.1587}$} & \it {\color[HTML]{000000}$\mathit{0.2390}$}  \\

        \midrule
        \multirow{10}{*}{SQuAD} 
        & Likelihood &  {\color[HTML]{000000}$0.3635$} & {\color[HTML]{000000}$0.6388$} & {\color[HTML]{000000}$0.5633$} & {\color[HTML]{000000}$0.6408$} & {\color[HTML]{000000}$0.5516$}  \\
        & Entropy & {\color[HTML]{000000}$0.6931$ } & {\color[HTML]{000000}$0.5920$} & {\color[HTML]{000000}$0.5993$} & {\color[HTML]{000000}$0.5426$} & {\color[HTML]{000000}$0.6068$}  \\
        & LogRank & {\color[HTML]{000000}$0.3395$} & {\color[HTML]{000000}$0.6410$} & {\color[HTML]{000000}$0.5368$} & {\color[HTML]{000000}$0.6167$} & {\color[HTML]{000000}$0.5268$}  \\
        & LRR & {\color[HTML]{000000}$0.2996$} & {\color[HTML]{000000}$0.5150$} & {\color[HTML]{000000}$0.4524$} & {\color[HTML]{000000}$0.5291$} & {\color[HTML]{000000}$0.4490$} \\
        & \note{DNA-GPT $\diamondsuit$} & {\color[HTML]{000000}$0.4916$} & {\color[HTML]{000000}$0.6584$} & {\color[HTML]{000000}$0.6172$} & {\color[HTML]{000000}$0.6782$} & {\color[HTML]{000000}$0.6114$}  \\
        & NPR $\diamondsuit$ & {\color[HTML]{000000}$0.4399$} & {\color[HTML]{000000}$0.6511$} & {\color[HTML]{000000}$0.5479$} & {\color[HTML]{000000}$0.5449$} & {\color[HTML]{000000}$0.5460$}  \\
        & DetectGPT $\diamondsuit$ & {\color[HTML]{000000}$0.5396$}& {\color[HTML]{000000}$0.7229$} & {\color[HTML]{000000}$0.6410$} & {\color[HTML]{000000}$0.6320$} & {\color[HTML]{000000}$0.6339$}  \\
        & Fast-DetectGPT & \underline{\color[HTML]{000000}$0.7056$ } & \underline{\color[HTML]{000000}$0.8855$ } & \underline{\color[HTML]{000000}$0.8317$ } & \underline{\color[HTML]{000000}$0.8344$ } & \underline{\color[HTML]{000000}$0.8143$ }  \\
        \rowcolor[gray]{.9}
        & ImBD (Ours) & \bf{\color[HTML]{000000}\bm{$0.8860$}} & \bf{\color[HTML]{000000}\bm{$0.9508$}} & \bf{\color[HTML]{000000}\bm{$0.9136$}} & \bf{\color[HTML]{000000}\bm{$0.9161$}} &  \bf{\color[HTML]{000000}\bm{$0.9166$}}  \\
        \rowcolor[gray]{.9}
        & \it (Diff) & {\color[HTML]{000000} \it $\mathit{0.1804}$} & \it{\color[HTML]{000000}$\mathit{0.0653}$} & {\color[HTML]{000000} \it $\mathit{0.0819}$} & \it {\color[HTML]{000000}$\mathit{0.0817}$} & \it {\color[HTML]{000000}$\mathit{0.1023}$} \\
        \midrule
        \multirow{10}{*}{WritingPrompts} 
        & Likelihood & {\color[HTML]{000000}$0.6208$} & {\color[HTML]{000000}$0.8500$} & {\color[HTML]{000000}$0.7657$} & {\color[HTML]{000000}$0.8816$} & {\color[HTML]{000000}$0.7795$}  \\
        & Entropy & {\color[HTML]{000000}$0.5903$} & {\color[HTML]{000000}$0.4371$} & {\color[HTML]{000000}$0.4692$} & {\color[HTML]{000000}$0.2941$} & {\color[HTML]{000000}$0.4476$}  \\
        & LogRank & {\color[HTML]{000000}$0.5694$} & {\color[HTML]{000000}$0.8192$} & {\color[HTML]{000000}$0.7147$} & {\color[HTML]{000000}$0.8675$} & {\color[HTML]{000000}$0.7427$}  \\        
        & LRR & {\color[HTML]{000000}0.4203} & {\color[HTML]{000000}$0.6876$} & {\color[HTML]{000000}$0.5261$} & {\color[HTML]{000000}$0.7880$} & {\color[HTML]{000000}$0.6055$}  \\
        & \note{DNA-GPT $\diamondsuit$} & {\color[HTML]{000000}$0.6795$} & {\color[HTML]{000000}$0.8244$} & {\color[HTML]{000000}$0.7545$} & {\color[HTML]{000000}$0.8725$} & {\color[HTML]{000000}$0.7827$}  \\
        & NPR $\diamondsuit$ & {\color[HTML]{000000}$0.7870$} & {\color[HTML]{000000}$0.8904$} & {\color[HTML]{000000}$0.7890$} & {\color[HTML]{000000}$0.8727$} & {\color[HTML]{000000}$0.8348$}  \\
        & DetectGPT $\diamondsuit$ & {\color[HTML]{000000}$0.8298$} & {\color[HTML]{000000}$0.8984$} & {\color[HTML]{000000}$0.8589$} & {\color[HTML]{000000}$0.8987$} & {\color[HTML]{000000}$0.8715$}  \\
        & Fast-DetectGPT 
         & \underline{\color[HTML]{000000}$0.8967$ } & \underline{\color[HTML]{000000}$0.9562$ } &  \underline{\color[HTML]{000000}$0.9141 $} & \underline{\color[HTML]{000000}$0.9539$ } &  \underline{\color[HTML]{000000}$0.9302$ } \\
         \rowcolor[gray]{.9}
         & ImBD (Ours) & \bf{\color[HTML]{000000}\bm{$0.9653$}} & \bf{\color[HTML]{000000}\bm{$0.9908$}} &  \bf{\color[HTML]{000000}\bm{$0.9670$}} & \bf{\color[HTML]{000000}\bm{$0.9796$}} &  \bf{\color[HTML]{000000}\bm{$0.9757$}}  \\
         \rowcolor[gray]{.9}
         & \it (Diff) & \it {\color[HTML]{000000}$\mathit{0.0686}$} & \it {\color[HTML]{000000}$\mathit{0.0346}$} & \it {\color[HTML]{000000}$\mathit{0.0529}$} & \it {\color[HTML]{000000}$\mathit{0.0257}$} & \it {\color[HTML]{000000}$\mathit{0.0455}$}  \\
\hline

\hline

\hline

\hline
    \end{tabular}
    \caption{
    \textbf{Performance on \emph{open-source model} polished text.} AUROC scores are averaged across the datasets generated by the polish task based on XSum, SQuAD, and WritingPrompts.  
    NPR and DetectGPT use T5-3B/Neo-2.7 as the perturbation/scoring models and Fast-DetectGPT uses GPT-J/Neo-2.7 as the sampling/scoring models.
    An underline denotes the second-best AUROC. The ``{\it (Diff)}'' rows indicate the AUROC improvement upon the second-best baselines. $\diamondsuit$ -- Methods call models a hundred times, thus consuming much higher computational resources. ImBD is trained on $500$ sample pairs of \texttt{polish} task.
    }
    \label{tab:main_result_details_polish}
\end{table*}
\begin{table*}[t]
\small
\centering
\setlength{\tabcolsep}{7.5pt}
\renewcommand\arraystretch{1.2}
    \begin{tabular}{l|c@{\hspace{9pt}}c@{\hspace{9pt}}c@{\hspace{9pt}}c|c@{\hspace{9pt}}c@{\hspace{9pt}}c@{\hspace{9pt}}c}
\hline

\hline

\hline

\hline
        \multirow{2}{*}{\bf Method} & \multicolumn{3}{c}{\bf GPT-3.5} &\multirow{2}{*}{\bf Avg.} &\multicolumn{3}{c}{\bf GPT-4o}&\multirow{2}{*}{\bf Avg.} \\
        & XSum & Writing & PubMed &  & XSum & Writing & PubMed &  \\
\hline

\hline
        RoBERTa-base~\citep{liu2019roberta} & {\color[HTML]{000000} $0.4269$} & {\color[HTML]{000000} $0.4526$}& {\color[HTML]{000000} $0.4817$} & {\color[HTML]{000000} $0.4537$} & {\color[HTML]{000000} $0.4649$} & {\color[HTML]{000000} $0.5335$} & {\color[HTML]{000000} $0.4524$} & {\color[HTML]{000000} $0.4836$} \\
        RoBERTa-large~\citep{liu2019roberta} & {\color[HTML]{000000} $0.5548$} & {\color[HTML]{000000} $0.5476$} & {\color[HTML]{000000} $0.4676$} &{\color[HTML]{000000} $0.5233$} & {\color[HTML]{000000} $0.5325$} & {\color[HTML]{000000} $0.5107$} & {\color[HTML]{000000} $0.4824$} &{\color[HTML]{000000} $0.5085$} \\
        
\hline
        Likelihood~\citep{ippolito2020automatic}& {\color[HTML]{000000} $0.2774$} & {\color[HTML]{000000} $0.5448$} & {\color[HTML]{000000} $0.4480$} & {\color[HTML]{000000} $0.4234$} & {\color[HTML]{000000} $0.4290$} & {\color[HTML]{000000} $0.6834$} & {\color[HTML]{000000} $0.4955$} & {\color[HTML]{000000} $0.5360$} \\
        Entropy~\citep{gehrmann2019gltr} & {\color[HTML]{000000} $0.6236$} & {\color[HTML]{000000} $0.4563$} & {\color[HTML]{000000} $0.5160$} & {\color[HTML]{000000} $0.5320$} & {\color[HTML]{000000} $0.5351$} & {\color[HTML]{000000} $0.3281$} & {\color[HTML]{000000} $0.4923$} & {\color[HTML]{000000} $0.4518$} \\
        LogRank~\citep{solaiman2019release} & {\color[HTML]{000000} $0.2528$} & {\color[HTML]{000000} $0.4847$} & {\color[HTML]{000000} $0.4454$} & {\color[HTML]{000000} $0.3943$} & {\color[HTML]{000000} $0.4064$} & {\color[HTML]{000000} $0.6581$} & {\color[HTML]{000000} $0.4936$} & {\color[HTML]{000000} $0.5194$} \\
        LRR~\citep{su2023detectllm} & {\color[HTML]{000000} $0.2185$} & {\color[HTML]{000000} $0.3208$} & {\color[HTML]{000000} $0.4505$} & {\color[HTML]{000000} $0.3299$} & {\color[HTML]{000000} $0.3647$} & {\color[HTML]{000000} $0.5528$} & {\color[HTML]{000000} $0.4820$} & {\color[HTML]{000000} $0.4665$} \\
        \note{DNA-GPT~\citep{yang2023dna}} & {\color[HTML]{000000} $0.2720$} & {\color[HTML]{000000} $0.5170$} & {\color[HTML]{000000} $0.4020$} & {\color[HTML]{000000} $0.3970$} & {\color[HTML]{000000} $0.4258$} & {\color[HTML]{000000} $0.6006$} & {\color[HTML]{000000} $0.4688$} & {\color[HTML]{000000} $0.4984$} \\   
        NPR~\citep{su2023detectllm} & {\color[HTML]{000000}$0.2873$} &{\color[HTML]{000000}$0.5753$} & {\color[HTML]{000000}$0.4248$} &{\color[HTML]{000000}$0.4487$} &{\color[HTML]{000000}$0.4066$} & {\color[HTML]{000000}$0.7067$} &{\color[HTML]{000000}$0.4811$} & {\color[HTML]{000000}$0.5315$}\\
        DetectGPT~\citep{mitchell2023detectgpt} & {\color[HTML]{000000}$0.3118$} &{\color[HTML]{000000} $0.6023$} & {\color[HTML]{000000}$0.4320$} &{\color[HTML]{000000}$0.4487$} &{\color[HTML]{000000}$0.4350$} & {\color[HTML]{000000}$0.7270$} &{\color[HTML]{000000}$0.4949$} & {\color[HTML]{000000}$0.5523$}\\
        Fast-Detect~\cite{bao2023fast} & {\color[HTML]{000000} $0.2683$} & {\color[HTML]{000000} $0.5518$} & {\color[HTML]{000000} $0.4407$} & {\color[HTML]{000000} $0.4203$} & {\color[HTML]{000000} $0.3961$} & {\color[HTML]{000000} $0.6212$} & {\color[HTML]{000000} $0.4847$} & {\color[HTML]{000000} $0.5007$} \\
        \rowcolor[gray]{.9} 
        ImBD (Ours)& \bf {\color[HTML]{000000}\bm{$0.8651$}} & \bf{\color[HTML]{000000} \bm{$0.8828$}} & \bf{\color[HTML]{000000} \bm{$0.6218$}} & \bf{\color[HTML]{000000} \bm{$0.7899$}} & \bf{\color[HTML]{000000} \bm{$0.7995$}} & \bf{\color[HTML]{000000} \bm{$0.8136$}} & \bf{\color[HTML]{000000} \bm{$0.6178$}} & \bf{\color[HTML]{000000} \bm{$0.7436$}} \\
\hline

\hline

\hline

\hline
    \end{tabular}
        \caption{\textbf{Performance on detecting \emph{GPT-3.5} and \emph{GPT-4o} rewritten text}.
        Typically, the Neo-2.7B~\cite{black2021gpt} is used as the source for scoring model. NPR and DetectGPT, on the other hand, utilize T5-3B~\cite{chen2019semantically} for generating perturbations, whereas Fast-DetectGPT employs GPT-J~\cite{wang2021gpt} as a surrogate model to generate samples. ImBD is trained in $500$ pairs of \texttt{polish} task. 
}
    \label{tab:chatgpt_gpt4_rewritten_results}
\end{table*}
\begin{table*}[t]
    \centering\small
    \setlength{\tabcolsep}{7pt}
    \begin{tabular}{l@{\hspace{10.5pt}}l@{\hspace{10.5pt}}c@{\hspace{10.5pt}}c@{\hspace{10.5pt}}c@{\hspace{10.5pt}}c@{\hspace{10.5pt}}c}
\hline

\hline

\hline

\hline
        \multirow{2}{*}{\bf Dataset} & \multirow{2}{*}{\bf Method} & \multicolumn{4}{c}{\bf Source Model} & \multirow{2}{*}{\bf Avg.}\\
         &   &  Qwen2-7B &  Llama-3-8B &  Mistral-7B &  Deepseek-7B &\\    
        \midrule
        \multirow{10}{*}{XSum} 
        & Likelihood~\citep{ippolito2020automatic} & {\color[HTML]{000000}$0.2741$} & {\color[HTML]{000000}$0.5851$} & {\color[HTML]{000000}$0.3613$} & {\color[HTML]{000000}$0.5170$} & {\color[HTML]{000000}$0.4344$}  \\
        & Entropy~\citep{gehrmann2019gltr} & \underline{\color[HTML]{000000}$0.6396$ } & {\color[HTML]{000000}$0.5165$} & \underline{\color[HTML]{000000}$0.6028$ } & {\color[HTML]{000000}$0.5862$} & \underline{\color[HTML]{000000}$0.5863$ }  \\
        & LogRank~\citep{solaiman2019release} & {\color[HTML]{000000}$0.2564$} & {\color[HTML]{000000}$0.5589$} & {\color[HTML]{000000}$0.3399$} & {\color[HTML]{000000}$0.5053$} & {\color[HTML]{000000}$0.4151$}  \\
        & LRR~\citep{su2023detectllm} & {\color[HTML]{000000}$0.2376$} & {\color[HTML]{000000}$0.4905$} & {\color[HTML]{000000}$0.3071$} & {\color[HTML]{000000}$0.4742$} & {\color[HTML]{000000}$0.3774$}  \\
        & \note{DNA-GPT~\citep{yang2023dna} $\diamondsuit$} & {\color[HTML]{000000}$0.3255$} & {\color[HTML]{000000}$0.5441$} & {\color[HTML]{000000}$0.4006$} & {\color[HTML]{000000}$0.4928$} & {\color[HTML]{000000}$0.4408$}  \\
        & NPR~\citep{su2023detectllm} $\diamondsuit$ & {\color[HTML]{000000}$0.2443$} & {\color[HTML]{000000}$0.4986$} & {\color[HTML]{000000}$0.2888$} & {\color[HTML]{000000}$0.4380$} & {\color[HTML]{000000}$0.3674$}  \\
        & DetectGPT~\citep{mitchell2023detectgpt} $\diamondsuit$ & {\color[HTML]{000000}$0.2726$} & {\color[HTML]{000000}$0.5436$} & {\color[HTML]{000000}$0.3115$} & {\color[HTML]{000000}$0.4512$} & {\color[HTML]{000000}$0.3947$}  \\
        & Fast-DetectGPT~\citep{bao2023fast} & {\color[HTML]{000000}$0.2853$} & \underline{\color[HTML]{000000}$0.6911$ } & {\color[HTML]{000000}$0.3938$} & \underline{\color[HTML]{000000}$0.6647$ } & {\color[HTML]{000000}$0.5087$}  \\
        \rowcolor[gray]{.9}
        & ImBD (Ours) & \bf {\color[HTML]{000000} \bm{$0.8952$}} & \bf  {\color[HTML]{000000}\bm{$0.9710$}} & \bf {\color[HTML]{000000}\bm{$0.8348$}} & \bf {\color[HTML]{000000}\bm{$0.8739$}} & \bf {\color[HTML]{000000}\bm{$0.8937$}}  \\
        \rowcolor[gray]{.9}
        & \it (Diff) & \it {\color[HTML]{000000}$\mathit{0.2556}$} & \it {\color[HTML]{000000}$\mathit{0.2799}$} & \it {\color[HTML]{000000}$\mathit{0.2320}$} & \it {\color[HTML]{000000}$\mathit{0.2092}$} & \it {\color[HTML]{000000}$\mathit{0.3074}$}  \\
        
        \midrule
        \multirow{10}{*}{SQuAD} 
        & Likelihood &  {\color[HTML]{000000}$0.3657$} & {\color[HTML]{000000}$0.6584$} & {\color[HTML]{000000}$0.5017$} & {\color[HTML]{000000}$0.6540$} & {\color[HTML]{000000}$0.5450$}  \\
        & Entropy & \underline{\color[HTML]{000000}$0.5718$ } & {\color[HTML]{000000}$0.4639$} & {\color[HTML]{000000}$0.5128$} & {\color[HTML]{000000}$0.4514$} & {\color[HTML]{000000}$0.5000$}  \\
        & LogRank & {\color[HTML]{000000}$0.3401$} & {\color[HTML]{000000}$0.6380$} & {\color[HTML]{000000}$0.4843$} & {\color[HTML]{000000}$0.6432$} & {\color[HTML]{000000}$0.5264$}  \\
        & LRR & {\color[HTML]{000000}$0.2925$} & {\color[HTML]{000000}$0.5385$} & {\color[HTML]{000000}$0.4256$} & {\color[HTML]{000000}$0.5858$} & {\color[HTML]{000000}$0.4606$} \\
        & \note{DNA-GPT $\diamondsuit$} & {\color[HTML]{000000}$0.4577$} & {\color[HTML]{000000}$0.6116$} & \underline{\color[HTML]{000000}$0.5413$ } & {\color[HTML]{000000}$0.6085$} & {\color[HTML]{000000}$0.5548$}  \\
        & NPR $\diamondsuit$ & {\color[HTML]{000000}$0.3238$} & {\color[HTML]{000000}$0.5539$} & {\color[HTML]{000000}$0.4346$} & {\color[HTML]{000000}$0.5490$} & {\color[HTML]{000000}$0.4653$}  \\
        & DetectGPT $\diamondsuit$ & {\color[HTML]{000000}$0.3550$}& {\color[HTML]{000000}$0.6051$} & {\color[HTML]{000000}$0.4609$} & {\color[HTML]{000000}$0.5763$} & {\color[HTML]{000000}$0.4993$}  \\
        & Fast-DetectGPT & {\color[HTML]{000000}$0.3764$} & \underline{\color[HTML]{000000}$0.7425$ } & {\color[HTML]{000000}$0.5272$} & \underline{\color[HTML]{000000}$0.7313$} & \underline{\color[HTML]{000000}$0.5944$ }  \\
        \rowcolor[gray]{.9}
        & ImBD (Ours) & \bf{\color[HTML]{000000}\bm{$0.7874$}} & \bf{\color[HTML]{000000}\bm{$0.9089$}} & \bf{\color[HTML]{000000}\bm{$0.7683$}} & \bf{\color[HTML]{000000}\bm{$0.7716$}} &  \bf{\color[HTML]{000000}\bm{$0.8091$}}  \\
        \rowcolor[gray]{.9}
        & \it (Diff) & {\color[HTML]{000000} \it $\mathit{0.2156}$} & \it{\color[HTML]{000000}$\mathit{0.1664}$} & {\color[HTML]{000000} \it $\mathit{0.2270}$} & \it {\color[HTML]{000000}$\mathit{0.0403}$} & \it {\color[HTML]{000000}$\mathit{0.2147}$} \\
 
        \midrule
        \multirow{10}{*}{WritingPrompts} 
        & Likelihood & {\color[HTML]{000000}$0.4354$} & {\color[HTML]{000000}$0.8435$} & {\color[HTML]{000000}$0.5133$} & {\color[HTML]{000000}$0.7708$} & {\color[HTML]{000000}$0.6408$}  \\
        & Entropy & {\color[HTML]{000000}$0.6013$} & {\color[HTML]{000000}$0.3442$} & {\color[HTML]{000000}$0.5440$} & {\color[HTML]{000000}$0.3579$} & {\color[HTML]{000000}$0.4619$}  \\
        & LogRank & {\color[HTML]{000000}$0.3810$} & {\color[HTML]{000000}$0.8068$} & {\color[HTML]{000000}$0.4640$} & {\color[HTML]{000000}$0.7466$} & {\color[HTML]{000000}$0.5996$}  \\        
        & LRR & {\color[HTML]{000000}$0.2457$} & {\color[HTML]{000000}$0.6418$} & {\color[HTML]{000000}$0.3282$} & {\color[HTML]{000000}$0.6494$} & {\color[HTML]{000000}$0.4663$}  \\
        & \note{DNA-GPT $\diamondsuit$} & {\color[HTML]{000000}$0.5860$} & {\color[HTML]{000000}$0.7808$} & {\color[HTML]{000000}$0.5801$} & {\color[HTML]{000000}$0.7260$} & {\color[HTML]{000000}$0.6682$}  \\
        & NPR $\diamondsuit$ & {\color[HTML]{000000}$0.5864$} & {\color[HTML]{000000}$0.8101$} & {\color[HTML]{000000}$0.5309$} & {\color[HTML]{000000}$0.7240$} & {\color[HTML]{000000}$0.6629$}  \\
        & DetectGPT $\diamondsuit$ & \underline{\color[HTML]{000000}0.6323 } & {\color[HTML]{000000}$0.8380$} & {\color[HTML]{000000}$0.5877$} & {\color[HTML]{000000}$0.7518$} & {\color[HTML]{000000}$0.7025$}  \\
        & Fast-DetectGPT &  {\color[HTML]{000000}$0.6089$}
         & \underline{\color[HTML]{000000}$0.9338$ } & \underline{\color[HTML]{000000}$0.6480$ } &  \underline{\color[HTML]{000000}$0.8408$ } & \underline{\color[HTML]{000000}$0.7579$ }  \\
         \rowcolor[gray]{.9}
         & ImBD (Ours) & \bf{\color[HTML]{000000}\bm{$0.8845$}} & \bf{\color[HTML]{000000}\bm{$0.9761$}} &  \bf{\color[HTML]{000000}\bm{$0.8384$}} & \bf{\color[HTML]{000000}\bm{$0.9020$}} &  \bf{\color[HTML]{000000}\bm{$0.9003$}}  \\
         \rowcolor[gray]{.9}
         & \it (Diff) & \it {\color[HTML]{000000}$\mathit{0.2522}$} & \it {\color[HTML]{000000}$\mathit{0.0423}$} & \it {\color[HTML]{000000}$\mathit{0.1904}$} & \it {\color[HTML]{000000}$\mathit{0.0612}$} & \it {\color[HTML]{000000}$\mathit{0.1424}$}  \\

\hline

\hline

\hline
    \end{tabular}
    \caption{\textbf{Details of detection on \emph{open-source model} rewritten text.} AUROC scores are averaged across the datasets generated by the rewrite task based on XSum, SQuAD, and WritingPrompts. NPR and DetectGPT use T5-3B/Neo-2.7 as the perturbation/scoring models and Fast-DetectGPT uses GPT-J/Neo-2.7 as the sampling/scoring models. 
    An underline denotes the second-best AUROC. The ``{\it (Diff)}" rows indicate the AUROC improvement upon the second-best baselines. $\diamondsuit$ -- Methods call models a hundred times, thus consuming much higher computational resources. ImBD is trained on $500$ sample pairs of \texttt{polish} task.}
    \label{tab:main_result_details_rewritten}
\end{table*}
\begin{table*}[t]
\setlength{\tabcolsep}{15pt}
    \centering\small
    \begin{tabular}{l@{\hspace{16.5pt}}l@{\hspace{16.5pt}}c@{\hspace{16.5pt}}c@{\hspace{16.5pt}}c@{\hspace{16.5pt}}c@{\hspace{16.5pt}}c}

\hline

\hline

\hline

\hline
        \multirow{2}{*}{\bf Model} & \multirow{2}{*}{\bf Method} & \multicolumn{4}{c}{\bf Tasks} & \multirow{2}{*}{\bf Avg.}\\
         &   &  Rewrite  &  Polish &  Expand &Generate  & \\    
        \midrule
        \multirow{10}{*}{ChatGPT} 
        & Likelihood~\citep{ippolito2020automatic} &{\color[HTML]{000000}$0.2774$} &{\color[HTML]{000000}$0.4982$} &{\color[HTML]{000000}$0.6105$} &{\color[HTML]{000000}$0.9577$} &{\color[HTML]{000000}$0.5860$}\\
        & Entropy~\citep{gehrmann2019gltr} &{\color[HTML]{000000}$0.6236$ } &{\color[HTML]{000000}$0.6742$} &{\color[HTML]{000000}$0.5390$} &{\color[HTML]{000000}$0.3305$} &{\color[HTML]{000000}$0.5418$}\\
        & LogRank~\citep{solaiman2019release} &{\color[HTML]{000000}$0.2528$} &{\color[HTML]{000000}$0.4711$} &{\color[HTML]{000000}$0.5849$} &{\color[HTML]{000000}$0.9583$} &{\color[HTML]{000000}$0.5668$}\\
        & LRR~\citep{su2023detectllm} &{\color[HTML]{000000}$0.2185$} &{\color[HTML]{000000}$0.4016$} &{\color[HTML]{000000}$0.5039$} &{\color[HTML]{000000}$0.9164$} &{\color[HTML]{000000}$0.5101$}\\
        & \note{DNA-GPT~\citep{yang2023dna}} &{\color[HTML]{000000}$0.2720$} &{\color[HTML]{000000}$0.5338$} &{\color[HTML]{000000}$0.5706$} &{\color[HTML]{000000}$0.9328$} &{\color[HTML]{000000}$0.5773$}\\
        & NPR~\citep{su2023detectllm} &{\color[HTML]{000000}$0.2873$} &{\color[HTML]{000000}$0.5659$} &{\color[HTML]{000000}$0.5856$} &{\color[HTML]{000000}$0.8587$} &{\color[HTML]{000000}$0.5744$}\\
        & DetectGPT~\citep{mitchell2023detectgpt} &{\color[HTML]{000000}$0.3118$} &{\color[HTML]{000000}$0.6343$} &{\color[HTML]{000000}$0.6564$} &{\color[HTML]{000000}$0.8796$} &{\color[HTML]{000000}$0.6201$}\\
        & Fast-DetectGPT~\citep{bao2023fast} &{\color[HTML]{000000}$0.2683$} &{\color[HTML]{000000}$0.7312$} &{\color[HTML]{000000}$0.7801$} &{\color[HTML]{000000}$0.9906$} &{\color[HTML]{000000}$0.6926$}\\
        \rowcolor[gray]{.9}
        & ImBD (Ours) & \bf {\color[HTML]{000000}\bm{$0.8651$}} & \bf  {\color[HTML]{000000}\bm{$0.9849$}} & \bf {\color[HTML]{000000}\bm{$0.9900$}} & \bf {\color[HTML]{000000}\bm{$0.9999$}} & \bf {\color[HTML]{000000}\bm{$0.9600$}}  \\

        \midrule
        \multirow{10}{*}{GPT-4o} 
        & Likelihood &{\color[HTML]{000000}$0.4290$} &{\color[HTML]{000000}$0.4396$} &{\color[HTML]{000000}$0.5333$} &{\color[HTML]{000000}$0.7585$} &{\color[HTML]{000000}$0.5401$}\\
        & Entropy &{\color[HTML]{000000}$0.5351$ } &{\color[HTML]{000000}$0.6122$} &{\color[HTML]{000000}$0.4867$} &{\color[HTML]{000000}$0.4792$} &{\color[HTML]{000000}$0.5283$}\\
        & LogRank &{\color[HTML]{000000}$0.4064$} &{\color[HTML]{000000}$0.4002$} &{\color[HTML]{000000}$0.5060$} &{\color[HTML]{000000}$0.7486$} &{\color[HTML]{000000}$0.5153$}\\
        & LRR &{\color[HTML]{000000}$0.3647$} &{\color[HTML]{000000}$0.3095$} &{\color[HTML]{000000}$0.4304$} &{\color[HTML]{000000}$0.6758$} &{\color[HTML]{000000}$0.4451$}\\
        & \note{DNA-GPT} &{\color[HTML]{000000}$0.4258$} &{\color[HTML]{000000}$0.4974$} &{\color[HTML]{000000}$0.5313$} &{\color[HTML]{000000}$0.7528$} &{\color[HTML]{000000}$0.5518$}\\
        & NPR &{\color[HTML]{000000}$0.4066$} &{\color[HTML]{000000}$0.5065$} &{\color[HTML]{000000}$0.5242$} &{\color[HTML]{000000}$0.7304$} &{\color[HTML]{000000}$0.5419$}\\
        & DetectGPT &{\color[HTML]{000000}$0.4350$} &{\color[HTML]{000000}$0.6217$} &{\color[HTML]{000000}$0.6318$} &{\color[HTML]{000000}$0.7928$} &{\color[HTML]{000000}$0.6203$} \\
        & Fast-DetectGPT &{\color[HTML]{000000}$0.3961$} &{\color[HTML]{000000}$0.6293$} &{\color[HTML]{000000}$0.6357$} &{\color[HTML]{000000}$0.8896$}  &{\color[HTML]{000000}$0.6377$}\\
        \rowcolor[gray]{.9}
        & ImBD (Ours) & \bf{\color[HTML]{000000}\bm{$0.7995$}} & \bf{\color[HTML]{000000}\bm{$0.9486$}} & \bf{\color[HTML]{000000}\bm{$0.9396$}} & \bf{\color[HTML]{000000}\bm{$0.9988$}} &  \bf{\color[HTML]{000000}\bm{$0.9216$}}  \\
  
        \midrule
        \multirow{10}{*}{Qwen2-7B} 
        & Likelihood &{\color[HTML]{000000} $0.2741$} &{\color[HTML]{000000} $0.2520$} &{\color[HTML]{000000} $0.3404$} & {\color[HTML]{000000} $0.7674$} & {\color[HTML]{000000} $0.4085$}\\
        & Entropy &{\color[HTML]{000000} $0.6396$ } &{\color[HTML]{000000} $0.7623$ } & {\color[HTML]{000000} $0.6823$ }& {\color[HTML]{000000} $0.5300$} & {\color[HTML]{000000} $0.6536$ }\\
        & LogRank &{\color[HTML]{000000} $0.2564$} &{\color[HTML]{000000} $0.2246$} &{\color[HTML]{000000} $0.3179$}& {\color[HTML]{000000} $0.7703$} & {\color[HTML]{000000} $0.3923$}\\        
        & LRR &{\color[HTML]{000000} $0.2376$}  & {\color[HTML]{000000} $0.1875$} &{\color[HTML]{000000} $0.2396$}& {\color[HTML]{000000} $0.7408$} & {\color[HTML]{000000} $0.3514$}\\
        & \note{DNA-GPT} &{\color[HTML]{000000} $0.3255$} &{\color[HTML]{000000} $0.3352$} & {\color[HTML]{000000} $0.3558$}& {\color[HTML]{000000} $0.7732$} & {\color[HTML]{000000} $0.4474$}\\
        & NPR &{\color[HTML]{000000} $0.2443$} & {\color[HTML]{000000} $0.3896$} &{\color[HTML]{000000} $0.3708$} & {\color[HTML]{000000} $0.7796$} & {\color[HTML]{000000} $0.4461$}\\
        & DetectGPT &{\color[HTML]{000000} $0.2726$}  &{\color[HTML]{000000} $0.4885$} &{\color[HTML]{000000} $0.4715$} & {\color[HTML]{000000} $0.8995$} & {\color[HTML]{000000} $0.5330$}\\
        & Fast-DetectGPT &{\color[HTML]{000000} $0.2853$} &{\color[HTML]{000000} $0.5945$}&{\color[HTML]{000000} $0.6000$} & {\color[HTML]{000000} $0.9625$ } & {\color[HTML]{000000} $0.6106$}\\
        \rowcolor[gray]{.9}
         & ImBD (Ours) & \bf{\color[HTML]{000000} \bm{$0.8952$}} & \bf{\color[HTML]{000000} \bm{$0.9589$}} &  \bf{\color[HTML]{000000} \bm{$0.9720$}} & \bf{\color[HTML]{000000} \bm{$1.0000$}} &  \bf{\color[HTML]{000000} \bm{$0.9565$}}  \\

        \midrule
        \multirow{10}{*}{Llama-3-8B} 
        & Likelihood &{\color[HTML]{000000} $0.5851$} &{\color[HTML]{000000} $0.5695$} &{\color[HTML]{000000} $0.6511$} &{\color[HTML]{000000} $0.9468$} &{\color[HTML]{000000} $0.6881$}\\
        & Entropy &{\color[HTML]{000000} 0.5165} &{\color[HTML]{000000} $0.6348$} &{\color[HTML]{000000} $0.6030$} &{\color[HTML]{000000} $0.4493$} &{\color[HTML]{000000} $0.5509$}\\
        & LogRank &{\color[HTML]{000000} $0.5589$} &{\color[HTML]{000000} $0.5412$} &{\color[HTML]{000000} $0.6447$} &{\color[HTML]{000000} $0.9499$} &{\color[HTML]{000000} $0.6739$}\\        
        & LRR &{\color[HTML]{000000} $0.4905$} & {\color[HTML]{000000} $0.4530$} &{\color[HTML]{000000} $0.5942$} &{\color[HTML]{000000} $0.9295$} &{\color[HTML]{000000} $0.6168$}\\
        & \note{DNA-GPT} &{\color[HTML]{000000} $0.5441$} &{\color[HTML]{000000} $0.5599$} &{\color[HTML]{000000} $0.6507$} &{\color[HTML]{000000} $0.9808$} &{\color[HTML]{000000} $0.6839$}\\
        & NPR &{\color[HTML]{000000} $0.4986$} & {\color[HTML]{000000} $0.6144$} &{\color[HTML]{000000} $0.6720$} &{\color[HTML]{000000} $0.9000$} &{\color[HTML]{000000} $0.6713$}\\
        & DetectGPT &{\color[HTML]{000000} $0.6536$} &{\color[HTML]{000000} $0.6904$} &{\color[HTML]{000000} $0.7632$} &{\color[HTML]{000000} $0.8891$} &{\color[HTML]{000000} $0.7491$}\\
        & Fast-DetectGPT &{\color[HTML]{000000} $0.6911$}& {\color[HTML]{000000} $0.8192$}&{\color[HTML]{000000} $0.9330$}&{\color[HTML]{000000} $0.9828$}&{\color[HTML]{000000} $0.8565$} \\
        \rowcolor[gray]{.9}
         & ImBD (Ours) & \bf{\color[HTML]{000000} \bm{$0.9710$}} & \bf{\color[HTML]{000000} \bm{$0.9884$}} &  \bf{\color[HTML]{000000} \bm{$0.9821$}} & \bf{\color[HTML]{000000} \bm{$0.9989$}} &  \bf{\color[HTML]{000000}\bm{$0.9851$}}  \\

        \midrule
        \multirow{10}{*}{Mistral-7B} 
        & Likelihood &{\color[HTML]{000000} $0.3613$} &{\color[HTML]{000000} $0.4353$} & {\color[HTML]{000000} $0.7056$} &{\color[HTML]{000000} $0.9443$} &{\color[HTML]{000000} $0.6116$}\\
        & Entropy &{\color[HTML]{000000} $0.6028$ } &{\color[HTML]{000000} $0.6539$} & {\color[HTML]{000000} $0.4864$} &{\color[HTML]{000000} $0.4059$} &{\color[HTML]{000000} $0.5373$}\\\
        & LogRank &{\color[HTML]{000000} $0.3399$} &{\color[HTML]{000000} $0.3980$} & {\color[HTML]{000000} $0.6828$} &{\color[HTML]{000000} $0.9406$} &{\color[HTML]{000000} $0.5903$}\\\    
        & LRR &{\color[HTML]{000000} $0.3071$} & {\color[HTML]{000000} $0.3112$} & {\color[HTML]{000000} $0.5985$} &{\color[HTML]{000000} $0.8924$} &{\color[HTML]{000000} $0.5273$}\\\
        & \note{DNA-GPT} &{\color[HTML]{000000} $0.4006$} &{\color[HTML]{000000} $0.4555$} & {\color[HTML]{000000} $0.6705$} &{\color[HTML]{000000} $0.9353$} &{\color[HTML]{000000} $0.6155$}\\\
        & NPR &{\color[HTML]{000000} $0.2888$} &{\color[HTML]{000000} $0.4594$}& {\color[HTML]{000000} $0.5885$} &{\color[HTML]{000000} $0.9096$} &{\color[HTML]{000000} $0.5616$}\\\
        & DetectGPT &{\color[HTML]{000000} $0.3115$} & {\color[HTML]{000000} $0.5480$} & {\color[HTML]{000000} $0.6964$} &{\color[HTML]{000000} $0.9573$} &{\color[HTML]{000000} $0.6283$}\\\
        & Fast-DetectGPT &{\color[HTML]{000000} $0.3938$} & {\color[HTML]{000000} $0.7034$} & {\color[HTML]{000000} $0.9161$} &{\color[HTML]{000000} $0.9984$} &{\color[HTML]{000000} $0.7529$}\\
        \rowcolor[gray]{.9}
         & ImBD (Ours) & \bf{\color[HTML]{000000} \bm{$0.8384$}} & \bf{\color[HTML]{000000} \bm{$0.9671$}} &  \bf{\color[HTML]{000000} \bm{$0.9946$}} & \bf{\color[HTML]{000000} \bm{$1.0000$}} &  \bf{\color[HTML]{000000} \bm{$0.9500$}}  \\

         \midrule
        \multirow{10}{*}{Deepseek-7B} 
        & Likelihood &{\color[HTML]{000000} $0.5170$} &{\color[HTML]{000000} $0.5438$} & {\color[HTML]{000000} $0.7822$}  & {\color[HTML]{000000} $0.9886$} & {\color[HTML]{000000} $0.7079$}\\
        & Entropy &{\color[HTML]{000000} $0.5862$} &{\color[HTML]{000000} $0.6402$} &{\color[HTML]{000000} $0.4609$} & {\color[HTML]{000000} $0.2822$} & {\color[HTML]{000000} $0.4924$}\\
        & LogRank &{\color[HTML]{000000} $0.5053$} &{\color[HTML]{000000}$0.5288$} & {\color[HTML]{000000} $0.7818$} & {\color[HTML]{000000} $0.9875$} & {\color[HTML]{000000} $0.7009$}\\        
        & LRR &{\color[HTML]{000000} $0.4742$} & {\color[HTML]{000000} $0.4859$} &{\color[HTML]{000000} $0.7431$} & {\color[HTML]{000000} $0.9696$} & {\color[HTML]{000000} $0.6682$}\\
        & \note{DNA-GPT} &{\color[HTML]{000000} $0.4928$} &{\color[HTML]{000000} $0.5586$} &{\color[HTML]{000000} $0.7295$} & {\color[HTML]{000000} $0.9837$} & {\color[HTML]{000000} $0.6912$}\\
        & NPR &{\color[HTML]{000000} $0.4380$} &{\color[HTML]{000000} $0.5476$} &{\color[HTML]{000000} $0.6628$} & {\color[HTML]{000000} $0.9465$} & {\color[HTML]{000000} $0.6487$}\\
        & DetectGPT &{\color[HTML]{000000} $0.4512$} &{\color[HTML]{000000} $0.6172$} &{\color[HTML]{000000} $0.7499$} & {\color[HTML]{000000} $0.9744$} & {\color[HTML]{000000} $0.6982$}\\
        & Fast-DetectGPT &{\color[HTML]{000000} $0.6647$} &{\color[HTML]{000000} $0.8177$} &{\color[HTML]{000000} $0.9286$} & {\color[HTML]{000000} $0.9996$} & {\color[HTML]{000000} $0.8527$}\\
        \rowcolor[gray]{.9}
         & ImBD (Ours) & \bf{\color[HTML]{000000} \bm{$0.8739$}} & \bf{\color[HTML]{000000} \bm{$0.9764$}} &  \bf{\color[HTML]{000000} \bm{$0.9766$}} & \bf{\color[HTML]{000000} \bm{$1.0000$}} &  \bf{\color[HTML]{000000} \bm{$0.9567$}}  \\
\hline

\hline

\hline

\hline
    \end{tabular}
  
  \caption{\textbf{Detail results across diverse machine text revision tasks on XSum.} NPR and DetectGPT use T5-3B/Neo-2.7 as the perturbation/scoring models and Fast-DetectGPT uses GPT-J/Neo-2.7 as the sampling/scoring models. ImBD is trained on $500$ sample pairs of \texttt{polish}
 task.}  \label{tab:domain_generlization_details}
\end{table*}


\end{document}